\documentclass[letterpaper]{article}
\usepackage{pub} 
\usepackage{times} 
\usepackage{helvet} 
\usepackage{courier} 
\usepackage[hyphens]{url} 
\usepackage{graphicx} 
\usepackage{graphicx} 
\usepackage{natbib} 
\usepackage{caption} 
\usepackage[utf8]{inputenc}  
\usepackage[polish]{babel} 

\usepackage{amsmath}
\usepackage{subcaption} 
\usepackage{lipsum} 
\usepackage{float}
\usepackage[T1]{fontenc}

\usepackage{xr}

\makeatletter
\newcommand*{\addFileDependency}[1]{
  \typeout{(#1)}
  \@addtofilelist{#1}
  \IfFileExists{#1}{}{\typeout{No file #1.}}
}
\makeatother

\newcommand*{\myexternaldocument}[1]{%
    \externaldocument{#1}%
    \addFileDependency{#1.tex}%
    \addFileDependency{#1.aux}%
}

\myexternaldocument{technical_appendix}

\title{PUB: Plot Understanding Benchmark and Dataset for Evaluating Large Language Models on Synthetic Visual Data Interpretation}

 \author{
 Aneta Pawelec\equalcontrib\textsuperscript{\rm 1, \rm 2},
 Victoria Sara Wesołowska\equalcontrib\textsuperscript{\rm 1, \rm 3},
 Zuzanna Bączek\equalcontrib\textsuperscript{\rm 1, \rm 3}, 
 Piotr Sankowski\textsuperscript{\rm 1, \rm 3}\\
 }
 \affiliations {
  \textsuperscript{\rm 1}Ideas NCBR\\
  \textsuperscript{\rm 2}Lodz University of Technology\\
  \textsuperscript{\rm 3}University of Warsaw
 }
 
\begin{document}

\maketitle

\begin{abstract}

  The ability of large language models (LLMs) to interpret visual representations of data is crucial for advancing their application in data analysis and decision-making processes. This paper presents a novel synthetic dataset designed to evaluate the proficiency of LLMs in interpreting various forms of data visualizations, including plots like time series, histograms, violins, boxplots, and clusters. Our dataset is generated using controlled parameters to ensure comprehensive coverage of potential real-world scenarios. We employ multimodal text prompts with questions related to visual data in images to benchmark several state-of-the-art models like ChatGPT or Gemini, assessing their understanding and interpretative accuracy.
  To ensure data integrity, our benchmark dataset is generated automatically, making it entirely new and free from prior exposure to the models being tested. This strategy allows us to evaluate the models' ability to truly interpret and understand the data, eliminating possibility of pre-learned responses, and allowing for an unbiased evaluation of the models' capabilities. We also introduce quantitative metrics to assess the performance of the models, providing a robust and comprehensive evaluation tool.
  Benchmarking several state-of-the-art LLMs with this dataset reveals varying degrees of success, highlighting specific strengths and weaknesses in interpreting diverse types of visual data. The results provide valuable insights into the current capabilities of LLMs and identify key areas for improvement. This work establishes a foundational benchmark for future research and development aimed at enhancing the visual interpretative abilities of language models. In the future, improved LLMs with robust visual interpretation skills can significantly aid in automated data analysis, scientific research, educational tools, and business intelligence applications.\end{abstract}

\section{Introduction}

In recent years, large language models (LLMs) have demonstrated remarkable capabilities in understanding and generating human language \cite{yao2024survey}. These advancements have sparked significant interest in their potential applications across diverse domains, including natural language processing and automated reasoning. As LLMs evolve, there is increasing emphasis on their multimodal capabilities \cite{yin2023survey}, particularly their ability to interpret and analyze visual data representations. This encompasses interpreting series plots, clusters, histograms, boxplots, and violin plots—areas where current LLMs encounter notable challenges. Integrating visual and textual data remains a complex task, underscoring the need for further advancements to enhance the effectiveness of LLMs in comprehensive data analysis.

A significant concern in the development and evaluation of multimodal LLMs is data contamination. When a model is trained or evaluated on data that it has previously encountered, the results can be misleading. For instance, \cite{liu2023mmc} used data generated by GPT-4 from publicly available Internet sources, while \cite{ChartBench} relied on Kaggle data. Such practices can result in an overestimation of the true capabilities of the model and do not accurately reflect its generalization performance \cite{balloccu2024leak}. This issue compromises the reliability of the benchmarks and impedes research progress by presenting an inflated view of the performance of the model.

To address these challenges, we introduce the Plot Understanding Benchmark (PUB), a novel synthetic data set designed to evaluate the proficiency of LLMs in interpreting various forms of data visualization. Our dataset is generated using controlled parameters to ensure comprehensive coverage of potential real-world scenarios. This approach not only eliminates the risk of data contamination, but also ensures that the evaluation is based solely on the model’s ability to interpret and understand visual data rather than relying on previous knowledge, maintaining its validity over time and ensuring that future evaluations remain unbiased and reflective of real model capabilities.

The nature of our benchmark allows for quantifying the LLM's responses to various kinds of visualizations and assessing its performance in a nuanced way. By manipulating different parameters within the generated plots, such as axis scales, data density, color schemes, and overall plot shape, we can systematically control the visual inputs presented to the LLM. This parameterization enables us to identify which aspects of the visualizations most significantly impact the LLM's ability to interpret and respond accurately. Furthermore, it allows for a detailed analysis of the LLM's sensitivity to specific features, thus providing insights into the underlying mechanisms of visual data interpretation. Through this approach, we can pinpoint strengths and weaknesses in the LLM's performance, facilitating targeted improvements and advancements in the development of more robust and capable models. Additionally, this method allows us to observe the conditions under which the model is more likely to produce hallucinations, enabling a clearer understanding of the factors that influence the odds of such occurrences and paving the way for mitigating these issues in future iterations of LLMs.

Furthermore, we introduce quantitative measures for assessing the performance of models, which have been previously missing in this context. These measures provide a robust and accurate tool for evaluating model capabilities, ensuring that our assessments are grounded in objective metrics rather than subjective evaluations.

Our study benchmarks several state-of-the-art LLMs such as GPT-4 \cite{achiam2023gpt}, Gemini \cite{team2023gemini}, or Claude \cite{anthropic2024claude} revealing varying degrees of success in interpreting different types of visual data. The results highlight specific strengths and weaknesses, offering valuable insights into the current capabilities of LLMs and identifying key areas for future improvement.

The implications of our findings are far-reaching. Enhancing the visual interpretative abilities of LLMs can significantly advance automated data analysis, scientific research, educational tools, and business intelligence applications. As such, this work not only provides a foundational benchmark for evaluating LLMs' visual interpretation skills but also lays the groundwork for future research and development aimed at creating more robust and versatile language models.

In the following sections, we detail the construction of our synthetic dataset, the methodology for benchmarking LLMs, the results of our evaluations, and the potential future applications of improved LLMs in various domains.

\section{Related Work}

Recent advancements in LLMs have demonstrated their impressive capabilities in understanding and generating human language. However, extending these abilities to multimodal tasks, particularly in visual data interpretation, presents unique challenges that have sparked considerable research interest.

\subsection{Multimodal Large Language Models}

Multimodal LLMs (MLLMs) aim to integrate text with visual or other non-textual data to enhance understanding and reasoning. For example, Li et al. \cite{li2024seedbench} introduced SEED-Bench, a benchmark designed to evaluate the proficiency of multimodal LLMs in processing and generating visual content. Their work highlights the importance of assessing LLMs' capabilities across diverse visual tasks, emphasizing the need for comprehensive benchmarks in this domain.

Similarly, Zhang et al. \cite{Zhang2023m3exam} proposed M3exam, a multilingual and multimodal benchmark to evaluate LLMs' performance across different tasks and languages, highlighting the versatility and potential of these models in handling complex, real-world scenarios.

\subsection{Benchmarking and Evaluation}

The development of benchmarks plays a crucial role in advancing the field by providing standardized methods to assess and compare the performance of different models. Chen et al. \cite{chen2024viseval} introduced VisEval, a benchmark specifically designed to evaluate LLMs in visual data analysis. Their work underscores the challenges LLMs face in interpreting complex visual representations and the importance of robust evaluation methodologies to ensure accurate performance assessments.

Further, Gadre et al. \cite{gadre2024datacomp} discussed the importance of dataset design in benchmarking, introducing DATACOMP as a new standard for multimodal dataset creation. This work highlights the significance of deduplication and data-centric approaches in enhancing the generalization capabilities of LLMs.

\subsection{Challenges in Visual Data Interpretation}

The ability of LLMs to accurately interpret and analyze visual data remains an underexplored area. Malode \cite{malode2024benchmarking} emphasized the need for optimized LLMs capable of handling multimodal data, identifying key challenges such as data contamination and overestimation of model performance due to exposure to training data during evaluation.

Moreover, McIntosh et al. \cite{mcintosh2024inadequacies} critically analyzed the inadequacies of current benchmarking practices, especially in the context of generative AI. Their work calls for a re-evaluation of existing benchmarks to better reflect the complex, multimodal nature of real-world tasks.

\section{Dataset Construction}

The objective of this paper is to benchmark the capability of multimodal models in understanding and interpreting various types of plotted data. To achieve this, we generate a variety of synthetic datasets, and create diverse visualizations of the results. This chapter details the steps involved in creating these datasets, ensuring they provide a comprehensive basis for evaluating the performance of multimodal models.

\subsection{Time Series}

\paragraph{Data Generation}

Our priority was to create artificial plots that resemble real-world data as closely as possible. To achieve this, we generate random time series data using a random walk process and a geometric random walk, as these methods are widely used in financial and economic data analysis \cite{Hull1989OptionsFA,Malkiel1973ARW}. The random walk process is defined as follows:

\begin{equation}
  W_t = \sum_{i=1}^{t} X_i
\end{equation}

where $X_i$ are independent and identically distributed random variables with a mean of zero and a standard deviation of one.

The geometric random walk is defined as follows:

\begin{equation}
  S_t = S_0 \text{exp} \left( \left( \mu - \frac{\sigma^2}{2} \right) t + \sigma W_t \right)
\end{equation}

where $S_0$ is the initial value, $\mu$ is the drift, $\sigma$ is the variance, and $W_t$ is the random walk process.

This process is parameterized by two additional values: drift and variance. The drift, $\mu$, represents the overall trend of the time series. A positive drift indicates an upward trend over time, suggesting consistent growth, whereas a negative drift implies a downward trend, indicating a gradual decline. The variance, $\sigma$, quantifies the degree of volatility or fluctuation in the time series. A higher variance means the data points are more spread out from the trend, leading to larger and more frequent changes, while a lower variance results in data points that are closer to the trend, producing smoother and more stable movements.

Having those two additional parameters allows us to measure how well the model responds to different kinds of plots and quantify its performance in a more nuanced way. The drift and variance values are randomly sampled from a predefined range to ensure a diverse set of time series plots with varying characteristics.

\paragraph{Data Transformation and Anomaly Introduction}  
To enhance the realism of artificial time series data and evaluate model robustness, we apply several data transformation techniques and introduce anomalies. Data smoothing is achieved using a moving average filter, where the window size is determined by a smoothing factor, \(0 < \text{factor} < 1\), to stabilize the data and aid trend identification. Pointwise anomalies simulate unexpected deviations at random points, with both the number and magnitude of anomalies being randomly determined to test the model's ability to detect and manage irregularities. To introduce variability in scale and offset, we randomize data ranges applying random shifts (\(\Delta_x, \Delta_y\)) and scaling factors (\(\text{scale}_x, \text{scale}_y\)), assessing the resilience of the model to changes in the scale and range of the data. In addition, consecutive data points are randomly removed to simulate missing-data scenarios, reflecting real-world issues such as sensor failures.

\subsection{Clusters}

\paragraph{Data Generation}
In order to check model ability to understand and interpret visualisation of clustered data we create diverse synthetic dataset by varying the number of clusters and samples. The number of clusters and the number of samples per cluster are randomly chosen to introduce variability. Cluster standard deviations are also randomized to ensure different degrees of overlap or separation between clusters. The data is generated using isotropic Gaussian blobs, and metadata such as cluster center location and standard deviations are recorded for each dataset for later evaluation.

\paragraph{Clustering Algorithms}
To ensure a comprehensive assessment, we apply a variety of clustering algorithms, randomly selecting from K-Means, Mean Shift, DBSCAN, or no clustering for each sample dataset. Different algorithms reveal diverse patterns and structures, providing a thorough evaluation of the model's capabilities. This approach tests the robustness of the model across various clustering scenarios and enhances its applicability to real-world data, where diverse clustering behaviors are common. Random selection ensures a broad range of clustering scenarios and, for algorithms requiring parameters, these are also randomly determined.

The clustering results are visualized using scatter plots with randomised options for marker styles, colors, and additional plot elements like legend.

\subsection{Histograms}
\paragraph{Data Generation}
To evaluate the models' ability to interpret histogram visualizations, we generate diverse synthetic datasets characterized by various parameters, including distribution type, size, and additional specifics such as mean, standard deviation, or skewness, to capture a wide range of real-world scenarios. The datasets are generated to follow several types of distributions, including uniform,   normal, exponential, Poisson, multimodal, and skewed (both left and right) distributions. This variety ensures that models encounter a broad spectrum of typical and complex statistical patterns. Parameters for each distribution type are randomly determined within realistic bounds to ensure variability. For example, the mean and standard deviation for normal distributions or the lambda for Poisson distributions are varied to create diverse datasets. Additionally, anomalies such as extra bins outside the normal range or the removal of certain bins are introduced randomly to test the models’ robustness and ability to detect irregular patterns.

\subsection{Boxplots and Violin Plots}
\paragraph{Data Generation}
We generate synthetic datasets to evaluate models on both boxplots and violin plots, using a variety of statistical distributions. For boxplots, data includes normal, log-normal, exponential, and mixed distributions, with varied parameters like mean, standard deviation, and scale.

For violin plots, we use normal, log-normal, exponential, gamma, beta, Weibull, Cauchy, uniform, and triangular distributions. Datasets feature 5-10 series and 50-100 points, with randomized parameters to simulate diverse real-world scenarios.

\subsection{Visualisation}
To create a more diverse dataset, we employ various visualization techniques with randomized settings, such as line colors, marker styles, grid presence, and axis scaling. This randomization introduces variability, ensuring that the models are tested under a wide range of visual conditions.

For time series, the line colors and the visibility of the grid are chosen randomly. Clustering results are visualized using scatter plots with various markers styles and plot elements. Histograms are created with different bin counts and visual settings like color and grid lines. Boxplots and violin plots are adjusted in terms of series count, color schemes, and axis ranges to reflect the data's spread and outliers.

This approach provides a comprehensive framework for assessing how visual presentation impacts the models' ability to interpret data, ensuring that the evaluation covers a broad spectrum of visual scenarios.
\subsection{Image Degradation}

To evaluate the models' ability to accurately interpret images under different conditions, we introduce various distortions. After generating the initial dataset, a portion of it is selected and augmented using one of the following methods, resulting in a second, modified dataset.

\paragraph{Noise}

We add a noise following a normal distribution to the image. The resulting image is a mixture of the original plot and noise:

\begin{equation}
  \text{image}_{\text{noisy}} = (1 - \alpha)\cdot \text{image}_{\text{real}} + \alpha \cdot \text{noise}
\end{equation}

where $\alpha$ is the noise coefficient.

\paragraph{Rotate}

We introduce rotation to the image, with the angle of the rotation randomly selected from the range (-60, 60).

\paragraph{Image Overlay}

To further challenge the models' interpretative abilities, we introduce an image overlay augmentation. This process involves pasting smaller, distinct images onto the original images within the dataset. By introducing these overlays, we simulate real-world scenarios where visual data may include occlusions, distractions, or additional objects.

\section{Benchmark procedure}

\subsection{Prompts}

In our benchmark, we employ multimodal prompts to evaluate the interpretive capabilities of LLMs on visual data. These prompts consist of a textual question paired with an image of a data plot, which requires the model to analyze the visual information and provide a structured response.

For instance, a prompt may ask the model to identify the largest cluster within a scatter plot and respond with the coordinates of the bounding box in JSON format. Another example involves approximating a plot with a series of points, where the model must generate an ordered list of coordinates that approximate the visual data.

The structured responses are requested in a specific JSON format, ensuring consistency and enabling precise evaluation of the model's performance. By varying the types of questions and the corresponding visual data, we comprehensively assess the models' abilities to interpret and respond to different visual scenarios.

\subsection{Time Series}

\paragraph{Detecting Minimal and Maximal Values}

This test assesses the proficiency of LLMs in identifying the minimum and maximum values within a dataset. Given a plot, the model is tasked with pinpointing the intervals that contain these extreme values. The performance score is calculated using the following formula:

\begin{equation}
  \mathcal{M} = 1 - \frac{(\min_{\text{pred}} - \min_{\text{real}})^2 + (\max_{\text{pred}} - \max_{\text{real}})^2}{(\min_{\text{real}} - \bar{y}_{\text{real}})^2 + (\max_{\text{real}} - \bar{y}_{\text{real}})^2}
\end{equation}

Here, $\min_{\text{pred}}$ and $\max_{\text{pred}}$ represent the predicted minimum and maximum values, while $\min_{\text{real}}$ and $\max_{\text{real}}$ denote the actual minimum and maximum values. The score intuitively reaches 1 if the model accurately identifies the minimal and maximal points and drops to 0 if the predicted intervals encompass the entire plot. This evaluation provides a clear metric for determining the model's accuracy in recognizing critical data points, thereby contributing to a comprehensive understanding of its capabilities in visual data analysis.

\begin{figure*}[t]
  \begin{subfigure}[t]{0.3\textwidth}
    \centering
    \includegraphics[width=\textwidth]{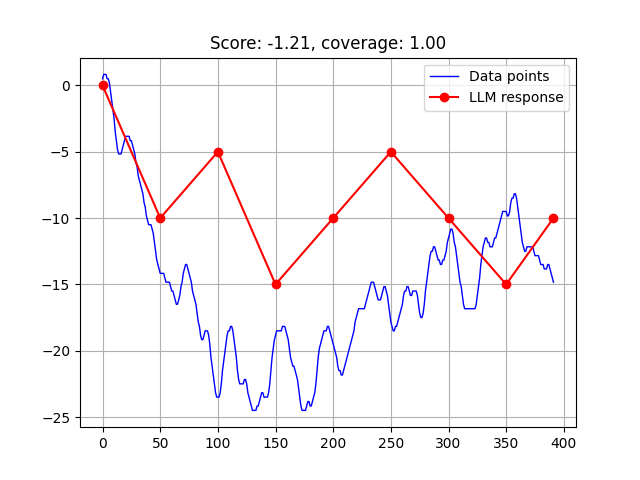}
    \caption{Geometric random walk with an answer to the approximate question.}
    \label{fig:geometric_walk_approximate}
  \end{subfigure}
    \hfill
      \centering
  \begin{subfigure}[t]{0.3\textwidth}
    \centering
    \includegraphics[width=\textwidth]{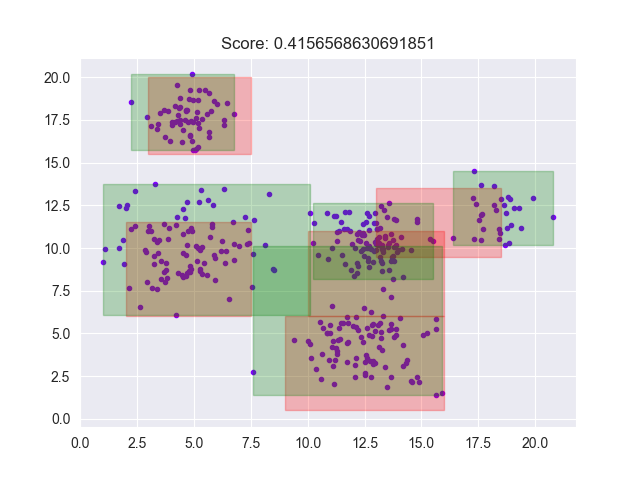}
    \caption{Clusters, without suggested clustering. In the green marked ground truth  answer to the question about clusters location and in red answer from gpt-4o.}
    \label{fig:clusters_area_GPT}
  \end{subfigure}
  \hfill
  \begin{subfigure}[t]{0.3\textwidth}
    \centering
    \includegraphics[width=\textwidth]{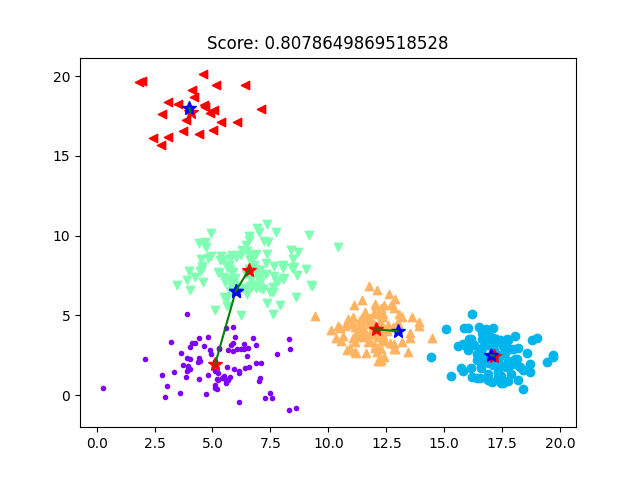} 
    \caption{Clusters with suggested clustering and an answer to the question about clusters' center location. Red star denotes ground truth and blue star marks the answer from gemini-1.5-flash.}
    \label{fig:centers_gemini_flash}
  \end{subfigure}
  \caption{Examples of plots with visualization of the answer from the model.}
  \label{fig:serie_full}
\end{figure*}

\paragraph{Data Approximation}
This test evaluates the LLM's ability to accurately interpret and replicate a given time series plot through a piecewise linear approximation (Figure \ref{fig:geometric_walk_approximate}). The LLM is presented with a time series plot and tasked with approximating it using up to nn points. The score is derived from the mean squared error between the LLM's piecewise linear approximation and the original plot.

\begin{gather}
  \mathcal{A} =
  1 -
  \frac{
  \sum_{x_i} (y_{\text{real}}(x_i) - y_{\text{approx}}(x_i))^2
  }{
  \sum_{x_i}(y_{\text{real}}(x_i) - \bar{y}_{\text{real}})^2
  }
\end{gather}

The piecewise linear approximation generated from the LLM's output serves as a robust indicator of the model's comprehension. High performance in this test suggests that the LLM can effectively detect and replicate trends, values, and overall patterns in the data, demonstrating a strong understanding of the graphical information presented. This makes our benchmark particularly valuable for assessing the practical capabilities of LLMs in interpreting visual data. Such an evaluation provides insights into the LLM's ability to process and approximate real-world data, highlighting its potential applicability in various analytical and decision-making tasks involving time series data.

\paragraph{Detecting Pointwise Anomalies}

We measure how well the model detects anomalies in the time series data. Given a plot, the model is tasked with identifying points that deviate significantly from the overall trend.

The LLM is asked to output the $x$ coordinate of the points it considers to be anomalies,
returning an empty list, should there be no anomalies.
To calculate the score, we check if the model predicted the correct number of anomalies
and how closely the predicted points match the actual anomalies.

If model predicts the correct number of anomalies, the score is calculated as follows:

\begin{equation}
  \mathcal{P}_{A} = 1 -\frac{ \sum_{a \in A} \min_{p \in P} \left ( a - p \right ) ^2}{\sum_{a} (a - \bar{x}) ^2 }
\end{equation}

Where $A$ is the set of actual anomalies, $P$ is the set of predicted anomalies, and $\bar{x}$ is the middle of the x-axis.

Intuitively, the score will be 1 if the model correctly identifies the anomalies, 0 if the model's guess is as good as just guessing the middle of the x-axis, and negative if it's worse.

\paragraph{Detecting Missing Points}

We evaluated the model's ability to detect missing points in the time series data. During the data generation process, we randomly remove a subset of points from the time series plot, covering a predefined percentage of the data.

Given a plot, the model is asked to identify the range where the points are missing, if there is any. To calculate the score, we first check if the model correctly identified the presence of missing points. Then we measure how closely the predicted range matches the actual range where the points are missing.

If the model correctly identifies the presence of missing points, the score is calculated as follows, using the Jacaard similarity coefficient:

\begin{equation}
  \mathcal{P}_{M} = \frac{|M \cap P|}{|M \cup P|}
\end{equation}

Where $M$ is the interval where the points are missing and $P$ is the interval predicted by the model.

\subsection{Clusters}

\paragraph{Detecting Clusters} The task presented to the LLMs is to determine the locations of clusters within a scatter plot and provide the coordinates of the bounding areas occupied by these clusters (Figure \ref{fig:clusters_area_GPT}).

The LLMs' responses are evaluated based on the accuracy of the detected clusters' locations. The primary metric used for this evaluation is the Intersection over Union (IoU). IoU is a standard metric in object detection and image segmentation tasks, it is defined as the area of the intersection divided by the area of the union of the two bounding boxes.

Intersection over Union (IoU) Calculation:

\begin{equation}
{\text {IoU}
=
  \frac{
  {\text {Area of Intersection}}}{
  {\text {Area of Union}}}
}
\end{equation}

The IoU metric provides a robust measure of how well the predicted cluster locations match the actual cluster locations. An IoU score of 1 indicates perfect alignment, whereas a score of 0 indicates no overlap. For each cluster, we calculate the IoU between the model-predicted bounding box and the ground truth bounding box. The overall performance is then assessed by averaging the IoU scores across all clusters.

\paragraph{Detecting Cluster's Center} This task involves identifying the centers of clusters within a scatter plot and providing the coordinates of these centers (Figure \ref{fig:centers_gemini_flash}).

To evaluate the models' performance in identifying cluster centers, we use a metric based on Euclidean distance.
This metric involves pairing the predicted cluster centers (p) with the ground truth (gt), so that each point of both groups is in at least one pair (m pairs). We compute the Euclidean distance between each paired ground truth and the predicted cluster center $\rVert gt_i - p_i \rVert$. This gives us a measure of how close the predictions are to the actual centers.

Additionally, we compute the Euclidean distance from each ground truth cluster center to the center of the dataset plot (c) $\rVert gt_i - c \rVert$ (for n clusters). This provides a baseline measure of the distribution of the clusters within the plot.

\begin{gather}
  \mathcal{D} =
  \frac{
  \sum_{i=1} ^{n} \rVert gt_i - c \rVert
  -
  \sum_{i=1} ^{m} \rVert gt_i - p_i \rVert
  }{
  \sum_{i=1} ^{n} \rVert gt_i - c \rVert
  }
\end{gather}

This evaluation metric ranges from $-\infty$ to 1, where a positive score indicates that the predicted cluster centers are closer to the ground truth centers than the ground truth centers are to the center of the dataset plot, suggesting high precision in the model's predictions. A score of zero implies that the model's predictions are, on average, as accurate as simply guessing the center of the plot. Conversely, a negative score indicates that the predicted centers are farther from the ground truth than the ground truth centers are from the plot center, highlighting significant inaccuracies in the model's predictions.

\subsection{Biggest Cluster}

To assess the model's ability to localize the largest cluster within the data, we evaluate three key aspects: the proportion of correctly enclosed points, the area efficiency of the bounding rectangle, and a penalty for incorrectly included points. These factors are combined into a final score, providing a comprehensive measure of the model's performance.

The final score \( S \) for cluster localization is calculated as the arithmetic mean of three components:

\begin{equation}
S = \frac{P_{\text{correct}} + P_{\text{area}} + P_{\text{penalty}}}{3}
\end{equation}

Where:

- \( P_{\text{correct}} = \frac{N_{\text{correct}}}{N_{\text{total}}} \) is the proportion of correctly enclosed points.

- \( P_{\text{area}} = \frac{A_{\text{cluster}}}{A_{\text{rect}}} \) compares the area of the minimal bounding box of the cluster to the area of the predicted rectangle.

- \( P_{\text{penalty}} = 1 - \frac{N_{\text{incorrect}}}{N_{\text{all}}} \) accounts for incorrectly included points.

This score reflects the accuracy and efficiency of the model in cluster location.

\subsection{Histograms}

\paragraph{Distribution Identification}  This task evaluates the model's ability to identify the distribution type in a histogram. The DistributionChecker class compares the model's predicted distribution against the actual distribution type in the metadata. The model scores 1.0 if the predicted distribution matches the actual distribution. Additionally, if the actual distribution is skewed (left or right) and the predicted distribution is exponential, the model also scores 1.0. This scoring accounts for cases where skewed distributions are often approximated by exponential distributions. This approach provides a robust evaluation of the model's ability to recognize different statistical distributions, accommodating both exact matches and reasonable approximations.

\paragraph{MinMax Evaluation}This evaluation assesses the model's accuracy in predicting the minimum and maximum ranges within histogram data using the Jaccard Similarity Index. The index measures the overlap between the predicted and actual intervals:

\begin{equation}
\text{Jaccard Similarity} = \frac{\text{Intersection}(A, B)}{\text{Union}(A, B)}
\end{equation}
The Jaccard similarity is computed separately for the minimum and maximum intervals, and the final evaluation metric, the Overall Score, is the average of these two similarities:
\begin{equation}
\text{Overall Score} = \frac{J_{\text{min}} + J_{\text{max}}}{2}
\end{equation}
Here, \(J_{\text{min}}\) denotes the average Jaccard similarity for the minimum and maximum intervals, respectively. This metric comprehensively measures the model's ability to predict data ranges within histograms accurately.

\paragraph{Monotonicity Evaluation} This task assesses the model's ability to identify monotonic intervals (either increasing or decreasing) in histogram data. The score is determined by the ratio of correctly predicted monotonic intervals to the total number of predicted intervals:

\begin{equation}
\mathcal{M} = \frac{C_{\text{inc}} + C_{\text{dec}}}{T_{\text{pred}}}
\end{equation}
Here, \({C_{\text {inc}}}\) is the count of correctly predicted increasing intervals, \(C_{\text{dec}}\) is the count of correctly predicted decreasing intervals, and \(T_{\text{pred}}\) is the total number of predicted monotonic intervals. This score reflects the model's ability to detect and represent underlying patterns accurately.

\paragraph{BelowXValuePercentage}This class evaluates the model's accuracy in predicting the percentage of data points falling below a specified threshold. The evaluation metric is:

\begin{equation}
\text{Score} = 1 - \frac{\left|\text{Actual Percentage} - \text{Predicted Percentage}\right|}{100}
\end{equation}
The score reflects how close the predicted percentage is to the actual percentage, minimizing the absolute difference.

\paragraph{FindAnomaly} This class assesses the accuracy of predicting anomaly ranges within histogram data using the Weighted Jaccard Similarity, which considers a radius around both predicted and actual anomaly ranges:
\begin{equation}
\text{Weighted Jaccard Similarity} = \frac{|E_{\text{int}}|}{|E_{\text{uni}}|}
\end{equation}
In this formula,\( |E_{\text{int}}| \) is the size of the intersection of the extended ranges, and  \( |E_{\text{uni}}| \) is the size of the union of the extended ranges. This metric measures the overlap between predicted and actual anomaly ranges, accommodating minor deviations for a comprehensive assessment.

\begin{table*}[!t]
\centering
\begin{tabular}{|l|c|c|c|c|c|}
\hline
\textbf{Model} & \textbf{Clustering} & \textbf{Histograms} & \textbf{Series} & \textbf{Boxplots} & \textbf{Violins} \\
\hline
gpt-4o & 0.549 & \textbf{0.57} & \textbf{0.44} & 0.491 & 0.472 \\
gpt-4o-mini & 0.432 & 0.40 & 0.19 & 0.342 & 0.356 \\
claude-3-5-sonnet & \textbf{0.682} & 0.46 & -317.38 & 0.603 & \textbf{0.579} \\
claude-3-opus & 0.495 & 0.32 & -23.64 & - & - \\
claude-3-haiku & 0.392 & 0.33 & -18.36 & 0.308 & 0.289 \\
gemini-1.5-pro & 0.637 & 0.54 & -123.45 & \textbf{0.607} & 0.572 \\
gemini-1.5-flash & 0.566 & 0.44 & -6520.23 & 0.513 & 0.471 \\
\hline
\end{tabular}
\caption{Overall model performance across different plot categories.}
\label{tab:summary_table}
\end{table*}

\subsection{Boxplots and Violin Plots}

\paragraph{Highest and Lowest Median}
For both boxplots and violin plots, the median of each plot is calculated to evaluate the predictions of the highest and lowest medians. The indices corresponding to the plots with the highest and lowest medians are identified and compared to predicted values. Correct identification of these indices earns 0.5 points for each correct prediction (one for the highest and one for the lowest), with a maximum score of 1.0.

\paragraph{Biggest and Smallest Range}
The range of data in each plot is determined by calculating the difference between the maximum and minimum values. For both types of plot, the indices of the plots with the largest and smallest ranges are identified. Then these indices are compared with predicted values. Correctly predicting the index of the plot with the largest range earns 0.5 points, as does correctly predicting the smallest range, for a total possible score of 1.0.

\paragraph{Biggest and Smallest IQR}
The interquartile range (IQR), calculated as the difference between the 75th and 25th percentiles, is used to evaluate predictions regarding variability within each plot. The indices of the plots with the largest and smallest IQRs are identified and compared to the predicted indices. Correct predictions earn 0.5 points each, resulting in a maximum score of 1.0.

\section{Experiments}

To evaluate the performance of multimodal models in interpreting various types of plots, we conducted a series of experiments across different plot categories. These experiments were designed to assess how well the models handle diverse visual presentations and tasks.

\subsection{Experimental Setup}

\paragraph{Models and Data}
We evaluated several state-of-the-art multimodal models, including GPT-4o, GPT-4o-mini, Claude-3 (various versions), and Gemini-1.5 (various versions). Each model was tested on a comprehensive dataset of synthetic plots, which included scatter plots, histograms, time series, boxplots, and violin plots. Dataset were create in two version, one without any distortions and other with random augmentation. 

\paragraph{Reproducibility}
All experiments were conducted under consistent conditions, and model performance was evaluated based on the pre-defined metrics. Details of the experimental setup, including model configurations and data generation parameters, are provided in the Appendix.

\subsection{Results}

Table \ref{tab:summary_table} presents the overall performance of each model across different plot categories. \textit{claude-3-5-sonnet} leads with the highest scores in both clustering (0.682) and violin plots (0.579), while \textit{gemini-1.5-pro} excels in boxplots (0.607). In contrast, \textit{gpt-4o-mini} consistently shows lower performance across most categories. Detailed results for each model on specific metrics are provided in the Technical appendix.

\paragraph{Clustering}

The performance of various models were evaluated based on their ability to identify the biggest cluster, detect cluster centers, and estimate cluster areas. Among the models, \textit{claude-3-5-sonnet} achieved the highest overall score of 0.682, excelling in identifying the largest cluster and determining cluster centers. In contrast, \textit{gpt-4o-mini} showed the lowest performance with an overall score of 0.432, indicating room for improvement in cluster detection and center localization.

\paragraph{Histograms}

The models' abilities to interpret histogram data were evaluated using various metrics, including distribution detection, identification of minimum and maximum bin values, monotonicity analysis, and estimating the percentage of data below a specific threshold. Model \textit{gpt-4o} led in overall performance with a score of 0.57, performing particularly well in predicting the percentage of data below a specified value. On the other hand, \textit{claude-3-opus} had the lowest overall score of 0.32, struggling particularly with distribution detection and monotonicity assessment.

\paragraph{Series}

The models' performance in handling series data was evaluate on broad rage of tasks, such as identifying minimum and maximum intervals, approximating the plot with points, and detecting pointwise anomalies. Notably, most models struggled with the approximation task, with some models like \textit{claude-3-5-sonnet} and \textit{gemini-1.5-flash} showing negative overall scores due to significant deviations in approximations. \textit{gpt-4o} was the most consistent performer with an overall score of 0.44, indicating its relative robustness in series-related tasks.

\paragraph{Boxplots}

In the boxplot evaluation the models were assessed based on their ability to identify medians, overall ranges, and interquartile ranges (IQR). \textit{gemini-1.5-pro} performed the best with an overall score of 0.607, particularly excelling in detecting IQRs. \textit{gpt-4o-mini}, however, showed the lowest performance with an overall score of 0.342, indicating challenges in accurately identifying key boxplot features.

\paragraph{Violins}

Lastly, the models' performance was evaluated based on their ability to interpret the violin plots. \textit{claude-3-5-sonnet} again emerged as a top performer with an overall score of 0.579, showing strong results in estimating medians and overall ranges. \textit{gpt-4o-mini}, however, lagged with an overall score of 0.356, indicating difficulties in accurately interpreting the density and distribution characteristics of the violin plots.

\section{Conclusion}

This study provides a detailed evaluation of multimodal models' capabilities in interpreting various types of plots, including clustering results, histograms, time series, boxplots, and violin plots. Our findings indicate significant variability in model performance across different tasks and visualization settings. The introduction of specialized metrics for assessing model accuracy has highlighted both strengths and limitations in current models.

The results underscore the need for robust models that can handle diverse visual data effectively. Future research should focus on improving model performance across different plot types and visualization settings to enhance overall accuracy and reliability.

\bibliography{main}

\begin{thebibliography}{16}
\providecommand{\natexlab}[1]{#1}

\bibitem[{Achiam et~al.(2023)Achiam, Adler, Agarwal, Ahmad, Akkaya, Aleman, Almeida, Altenschmidt, Altman, Anadkat et~al.}]{achiam2023gpt}
Achiam, J.; Adler, S.; Agarwal, S.; Ahmad, L.; Akkaya, I.; Aleman, F.~L.; Almeida, D.; Altenschmidt, J.; Altman, S.; Anadkat, S.; et~al. 2023.
\newblock Gpt-4 technical report.
\newblock \emph{arXiv preprint arXiv:2303.08774}.

\bibitem[{Anthropic(2024)}]{anthropic2024claude}
Anthropic, A. 2024.
\newblock The claude 3 model family: Opus, sonnet, haiku.
\newblock \emph{Claude-3 Model Card}, 1.

\bibitem[{Balloccu et~al.(2024)Balloccu, Schmidtová, Lango, and Dušek}]{balloccu2024leak}
Balloccu, S.; Schmidtová, P.; Lango, M.; and Dušek, O. 2024.
\newblock Leak, Cheat, Repeat: Data Contamination and Evaluation Malpractices in Closed-Source LLMs.
\newblock arXiv:2402.03927.

\bibitem[{Chen et~al.(2024)Chen, Zhang, Xu, Ren, and Yang}]{chen2024viseval}
Chen, N.; Zhang, Y.; Xu, J.; Ren, K.; and Yang, Y. 2024.
\newblock VisEval: A Benchmark for Data Visualization in the Era of Large Language Models.
\newblock \emph{arXiv preprint arXiv:2407.00981}.

\bibitem[{Gadre et~al.(2023)Gadre, Ilharco, Fang, Hayase, Smyrnis, Nguyen, Marten et~al.}]{gadre2024datacomp}
Gadre, S.~Y.; Ilharco, G.; Fang, A.; Hayase, J.; Smyrnis, G.; Nguyen, T.; Marten, R.; et~al. 2023.
\newblock DataComp: In search of the next generation of multimodal datasets.
\newblock In \emph{Advances in Neural Information Processing Systems}, volume~36, 27092--27112. Curran Associates, Inc.

\bibitem[{Hull(1989)}]{Hull1989OptionsFA}
Hull, J. 1989.
\newblock Options, Futures, and Other Derivatives.

\bibitem[{Li et~al.(2024)Li, Ge, Ge, Wang, Wang, Zhang, and Shan}]{li2024seedbench}
Li, B.; Ge, Y.; Ge, Y.; Wang, G.; Wang, R.; Zhang, R.; and Shan, Y. 2024.
\newblock SEED-Bench: Benchmarking Multimodal Large Language Models.
\newblock In \emph{Proceedings of the IEEE/CVF Conference on Computer Vision and Pattern Recognition (CVPR)}, 13299--13308.

\bibitem[{Liu et~al.(2023)Liu, Wang, Yao, Chen, Song, Cho, Yacoob, and Yu}]{liu2023mmc}
Liu, F.; Wang, X.; Yao, W.; Chen, J.; Song, K.; Cho, S.; Yacoob, Y.; and Yu, D. 2023.
\newblock MMC: Advancing Multimodal Chart Understanding with Large-scale Instruction Tuning.
\newblock \emph{arXiv preprint arXiv:2311.10774}.

\bibitem[{Malkiel(1973)}]{Malkiel1973ARW}
Malkiel, B.~G. 1973.
\newblock A Random Walk Down Wall Street.

\bibitem[{Malode(2024)}]{malode2024benchmarking}
Malode, V.~M. 2024.
\newblock Benchmarking public large language models.
\newblock \emph{Thesis, HAW Hamburg}.

\bibitem[{McIntosh et~al.(2024)McIntosh, Susnjak, Liu, Watters, and Halgamuge}]{mcintosh2024inadequacies}
McIntosh, T.~R.; Susnjak, T.; Liu, T.; Watters, P.; and Halgamuge, M.~N. 2024.
\newblock Inadequacies of Large Language Model Benchmarks in the Era of Generative Artificial Intelligence.
\newblock arXiv:2402.09880.

\bibitem[{Team et~al.(2023)Team, Anil, Borgeaud, Wu, Alayrac, Yu, Soricut, Schalkwyk, Dai, Hauth et~al.}]{team2023gemini}
Team, G.; Anil, R.; Borgeaud, S.; Wu, Y.; Alayrac, J.-B.; Yu, J.; Soricut, R.; Schalkwyk, J.; Dai, A.~M.; Hauth, A.; et~al. 2023.
\newblock Gemini: a family of highly capable multimodal models.
\newblock \emph{arXiv preprint arXiv:2312.11805}.

\bibitem[{Xu et~al.(2023)Xu, Du, Qi, Xu, Yuan, and Guo}]{ChartBench}
Xu, Z.; Du, S.; Qi, Y.; Xu, C.; Yuan, C.; and Guo, J. 2023.
\newblock ChartBench: A Benchmark for Complex Visual Reasoning in Charts.
\newblock \emph{ArXiv}, abs/2312.15915.

\bibitem[{Yao et~al.(2024)Yao, Duan, Xu, Cai, Sun, and Zhang}]{yao2024survey}
Yao, Y.; Duan, J.; Xu, K.; Cai, Y.; Sun, Z.; and Zhang, Y. 2024.
\newblock A survey on large language model (llm) security and privacy: The good, the bad, and the ugly.
\newblock \emph{High-Confidence Computing}, 100211.

\bibitem[{Yin et~al.(2023)Yin, Fu, Zhao, Li, Sun, Xu, and Chen}]{yin2023survey}
Yin, S.; Fu, C.; Zhao, S.; Li, K.; Sun, X.; Xu, T.; and Chen, E. 2023.
\newblock A survey on multimodal large language models.
\newblock \emph{arXiv preprint arXiv:2306.13549}.

\bibitem[{Zhang et~al.(2023)Zhang, Aljunied, Gao, Chia, and Bing}]{Zhang2023m3exam}
Zhang, W.; Aljunied, M.; Gao, C.; Chia, Y.~K.; and Bing, L. 2023.
\newblock M3Exam: A Multilingual, Multimodal, Multilevel Benchmark for Examining Large Language Models.
\newblock In \emph{37th Conference on Neural Information Processing Systems (NeurIPS 2023) Track on Datasets and Benchmarks.}

\end{thebibliography}

\end{document}


\maketitle

\tableofcontents


\newpage



\part{Image Augmentation}
\begin{figure}[th]
  \centering
  \begin{subfigure}[t]{0.3\textwidth}
    \centering
    \includegraphics[width=\textwidth]{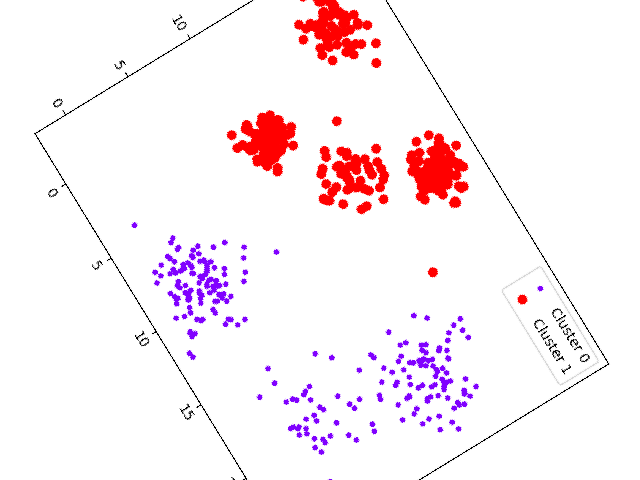}
    \caption{Example of image rotation. This transformation alters the orientation of the image.}
    \label{fig:rotated_image}
  \end{subfigure}
  \hfill
  \begin{subfigure}[t]{0.3\textwidth}
    \centering
    \includegraphics[width=\textwidth]{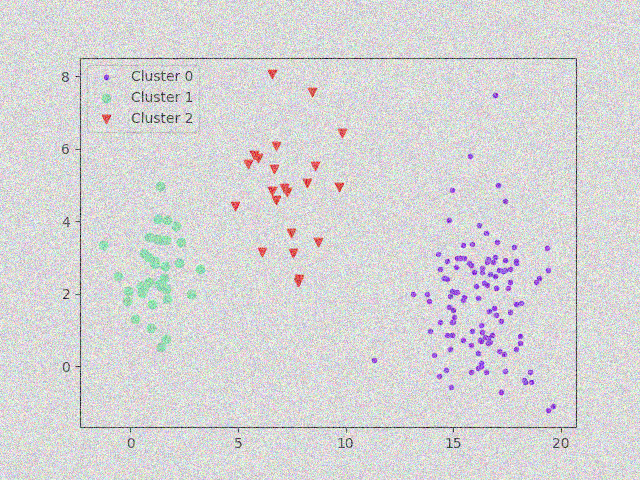}
    \caption{Example of image noise. This degradation adds random variations to pixel values.}
    \label{fig:image_noise}
  \end{subfigure}
  \hfill
  \begin{subfigure}[t]{0.3\textwidth}
    \centering
    \includegraphics[width=\textwidth]{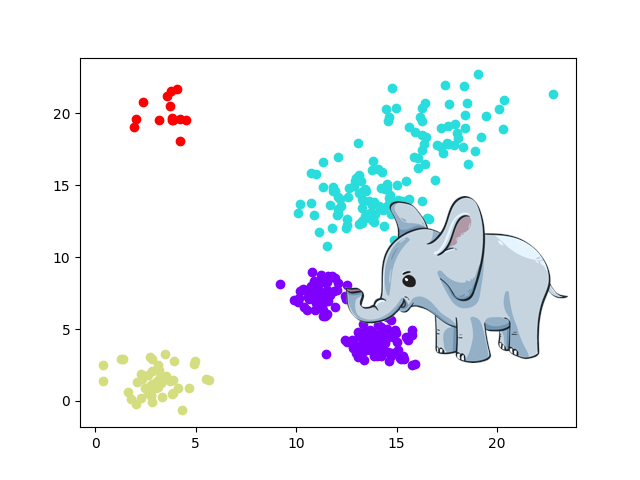}
    \caption{An example of the image overlay method, which combines data visualizations and object images into a single frame}
    \label{fig:image_overlay}
  \end{subfigure}
  \caption{Different types of image degradations as demonstrated in clusters. Each subfigure shows a specific type of image transformation: (a) rotation, (b) noise addition, and (c) overlay.}
  \label{fig:image_degradations}
\end{figure}


\begin{figure}[H]
  \begin{subfigure}[t]{0.3\textwidth}
    \centering
    \includegraphics[width=\textwidth]{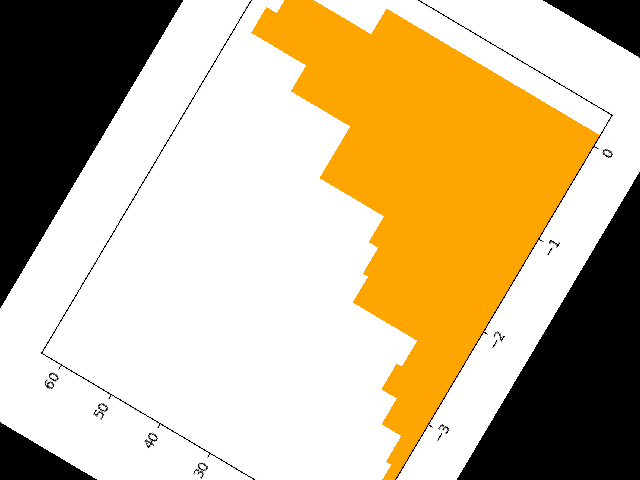}
    \caption{\tiny augmented 3aa1cf63-32cf-4037-91c9-bbc464876251}
  \end{subfigure}
    \hfill
      \centering
  \begin{subfigure}[t]{0.3\textwidth}
    \centering
    \includegraphics[width=\textwidth]{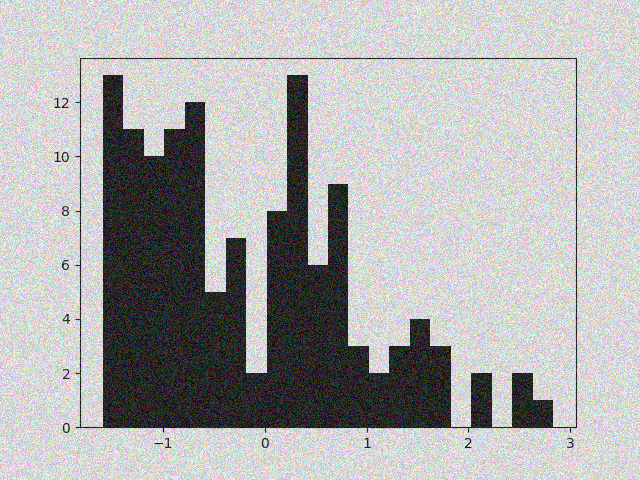}
    \caption{\tiny augmented 5a0d4464-2653-4914-a64a-9207308bc0c2}
  \end{subfigure}
  \hfill
  \begin{subfigure}[t]{0.3\textwidth}
    \centering
    \includegraphics[width=\textwidth]{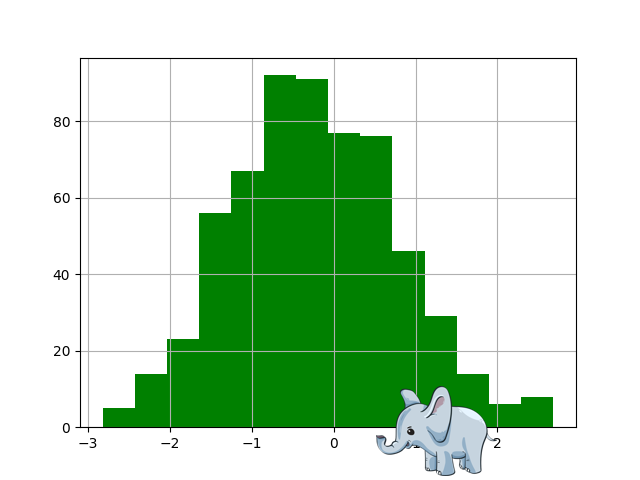} 
    \caption{\tiny augmented 0c7095b6-9877-4528-afcc-9b7359766e77}
  \end{subfigure}
  \hfill
  \begin{subfigure}[t]{0.3\textwidth}
    \centering
    \includegraphics[width=\textwidth]{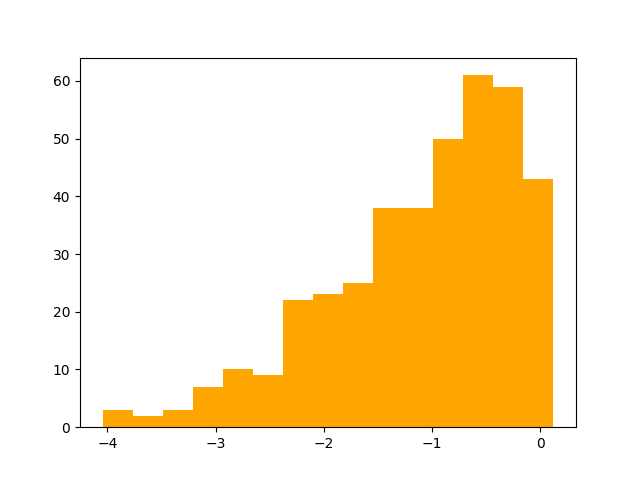} 
    \caption{\scriptsize 3aa1cf63-32cf-4037-91c9-bbc464876251}
  \end{subfigure}
  \hfill
  \begin{subfigure}[t]{0.3\textwidth}
    \centering
    \includegraphics[width=\textwidth]{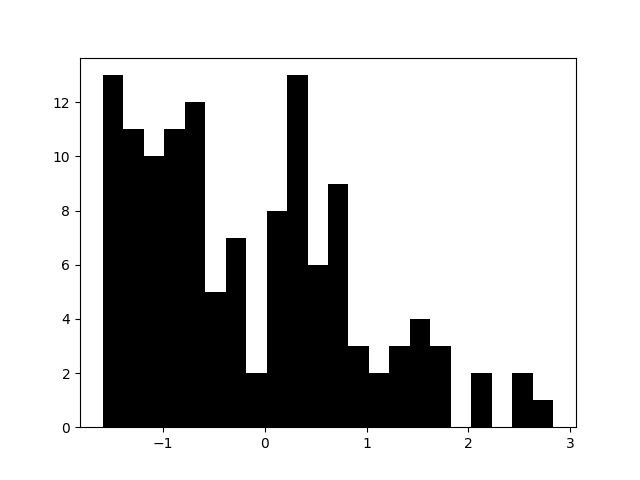} 
    \caption{\scriptsize 5a0d4464-2653-4914-a64a-9207308bc0c2}
  \end{subfigure}
  \hfill
  \begin{subfigure}[t]{0.3\textwidth}
    \centering
    \includegraphics[width=\textwidth]{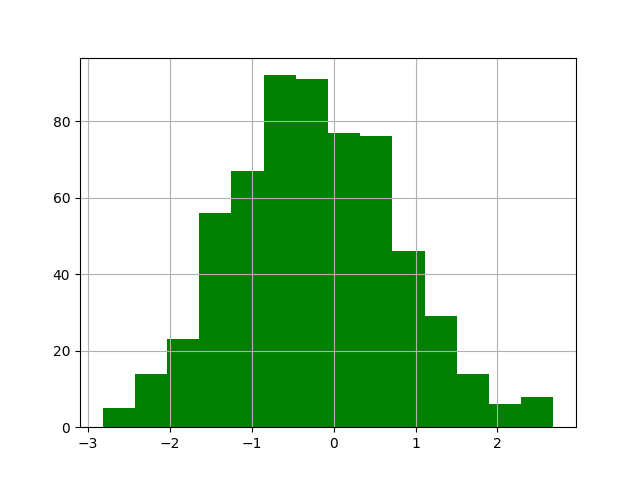} 
    \caption{\scriptsize 0c7095b6-9877-4528-afcc-9b7359766e77}
  \end{subfigure}
  \caption{Image Degradation vs Regular Images on histograms}
\end{figure}


\newpage
\part{Benchmarking procedure}
\section{Protocol}
\begin{figure}[H]
\begin{tikzpicture}[
    every node/.style={font=\sffamily},
    process/.style={rectangle, draw, rounded corners, thick, minimum width=2cm, minimum height=1cm, text centered, fill=blue!15, drop shadow},
    query/.style={rectangle, draw, rounded corners, thick, minimum width=2cm, minimum height=1cm, text centered, fill=green!15, drop shadow},
    response/.style={rectangle, draw, rounded corners, thick, minimum width=2cm, minimum height=1cm, text centered, fill=orange!15, drop shadow},
    eval/.style={rectangle, draw, rounded corners, thick, minimum width=5cm, minimum height=5cm, text centered, fill=purple!15, drop shadow},
    arrow/.style={-{Latex[length=2mm, width=1.5mm]}, thick, color=gray},
    dashedarrow/.style={dashed, gray, thick},
    node distance=1.5cm
    ]

    \node[process] (data) {
        \begin{minipage}[t]{3.2cm}
            \centering
            \textbf{Data Generation} \\[0.2cm]
             \includegraphics[width=3cm]{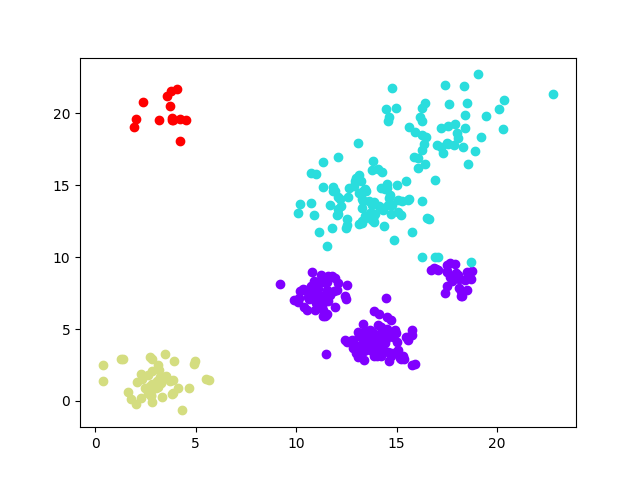} 
        \end{minipage}
    };
    
    \node[process, dashed] (augment) [right=2.5cm of data, anchor=center] {
        \begin{minipage}[t]{3.2cm}
            \centering
            \textbf{Image Augmentation} \\[0.2cm]
             \includegraphics[width=3cm]{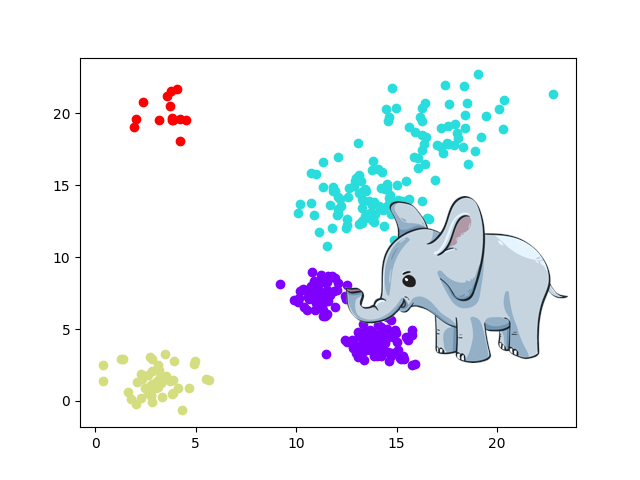} 
        \end{minipage}
    };

    \node (txt_prompt_img) [below=5cm of $(data)!0.5!(augment)$, anchor=center] {
        \includegraphics[width=8cm]{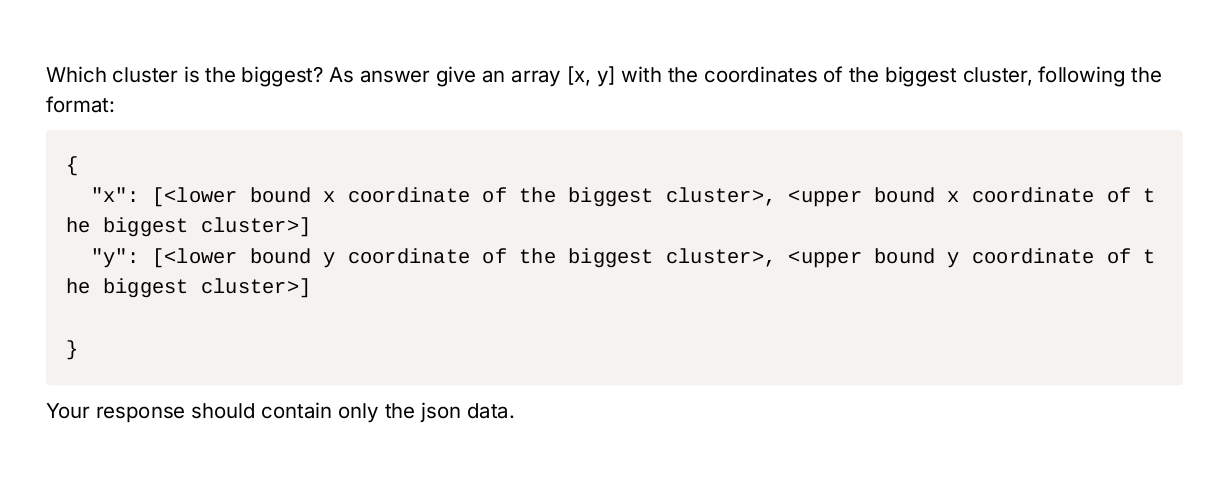}
    };

    \node[query, right=3.5cm of $(augment)!0.5!(txt_prompt_img)$] (query) {
        \textbf{MLLM}
    };

    \node[response, below=3cm of query] (response) {
        \textbf{Response}
    };

    \node[eval, right=1.5cm of response] (evaluation) {
        \begin{minipage}[t]{5.2cm}
            \centering
            \textbf{Evaluation} \\[0.2cm]
            \includegraphics[width=5cm]{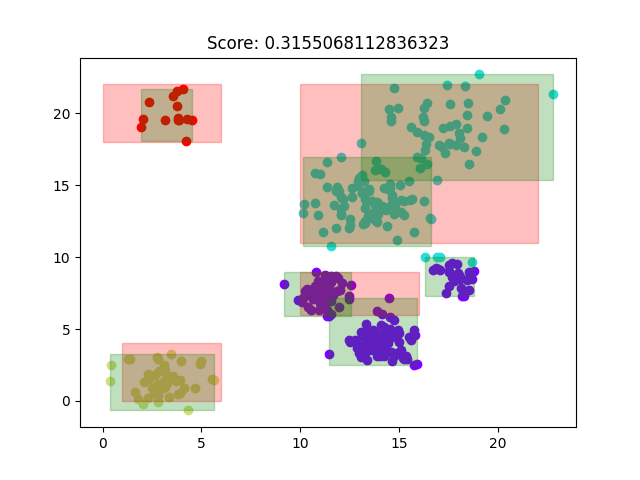}
        \end{minipage}
    };

    \draw[arrow] (data) -- (augment);
    \draw[arrow] (augment) -- (query);
    \draw[arrow] (txt_prompt_img) -- (query);
    \draw[arrow] (query) -- (response);
    \draw[arrow] (response) -- (evaluation);

    \node[below=0.35cm of data, align=center, font=\footnotesize] (img_input_label) {Sample Dataset};
    \node[below=0.3cm of augment, align=center, font=\footnotesize] (img_input_label) {Input Image};
    \node[below=0.3cm of txt_prompt_img, align=center, font=\footnotesize] (txt_prompt_img_label) {Text Prompt};

    \node[below=0.3cm of response, align=center, font=\footnotesize] (response_label) {Visualize Output};
    \node[below=0.3cm of evaluation, align=center, font=\footnotesize] (evaluation_label) {Analyze Output};

    \node [gray, above = 0.6cm of $(data.north west)!0.5!(augment.north east)$] {Multimodal Prompt Preparation};
    \node [gray, above = 0.6cm of $(query.north west)!0.5!(query.north east)$] {MLLM Query};
    \node [gray, above = 0.6cm of $(evaluation.north west)!0.5!(evaluation.north east)$] {Result Evaluation};

\end{tikzpicture}
\caption{Benchmark protocol.}
\label{fig:benchmark-protocol}
\end{figure}

Benchmark protocol involves preparing the input, querying a language model, obtaining and visualizing the response, and finally evaluating the results (Figure \ref{fig:benchmark-protocol}).

\section{Prompts}
\subsection{Prompt Design and Response Formatting}

In our benchmark, we carefully design multimodal prompts to evaluate the interpretive capabilities of MLLMs on visual data. Each prompt consists of a textual question paired with an image of a data plot, requiring the model to analyze the visual information and provide a structured response.

A key aspect of our prompt design is the inclusion of explicit instructions on how the response should be formatted. Specifically, models are instructed to return their answers directly in JSON format. This approach offers several advantages:

\begin{itemize}
    \item \textbf{Consistency}: By enforcing a structured response format, we ensure that all models provide their outputs in a uniform manner. This consistency is crucial for comparative evaluation across different models and tasks.
    
    \item \textbf{Ease of Parsing}: The use of JSON formatting simplifies the process of extracting and analyzing the model responses. Since JSON is a widely adopted data interchange format, it allows for straightforward parsing and further analysis, enabling precise measurement of model performance.
    
    \item \textbf{Precise Evaluation}: Structured responses in JSON format enable granular evaluation of the model's output. For example, if a model is asked to identify the largest cluster within a scatter plot, the prompt instructs it to return the coordinates of the bounding box in JSON. This not only allows us to verify the accuracy of the model's analysis but also facilitates automated scoring based on the expected structure.
\end{itemize}

By varying the types of questions and the corresponding visual data, we comprehensively assess the models' abilities to interpret and respond to different visual scenarios. The requirement for JSON-formatted responses further enhances the reliability of our benchmarking process, as it minimizes ambiguity in the model outputs and streamlines the evaluation workflow.

\subsection{Clusters}

\begin{figure}[H]
\begin{tcolorbox}[colback=lightgray, colframe=darkgray, title=biggest\_cluster.md]
Which cluster is the biggest?
As answer give an array [x, y] with the coordinates of the biggest cluster, following the format:

\begin{verbatim}
```json
{
  "x": [<lower bound x coordinate of the biggest cluster>,
    <upper bound x coordinate of the biggest cluster>]
    
  "y": [<lower bound y coordinate of the biggest cluster>, 
    <upper bound y coordinate of the biggest cluster>]

}
```
\end{verbatim}
Your response should contain only the json data.
\end{tcolorbox}
\caption{Prompt for identifying the biggest cluster in a dataset.}
    \label{fig:biggest_cluster}
\end{figure}

\begin{figure}[H]
    \centering
\begin{tcolorbox}[colback=lightgray, colframe=darkgray, title=centers.md]
Where are located the centers of the clusters?
As answer give an array [x, y] with the coordinates of for each cluster, following the format:

\begin{verbatim}
```json
{
  "<index>": [x, y],
}
```
\end{verbatim}
Your response should contain only the json data.
\end{tcolorbox}
\caption{Prompt for identifying the centers of clusters.}
    \label{fig:centers}
\end{figure}

\begin{figure}[H]
    \centering
\begin{tcolorbox}[colback=lightgray, colframe=darkgray, title=clusters\_area.md]
Where is each cluster located? Please provide the coordinates of the areas occupied by distinct clusters.
Give your answer based on points locations. Suggested clustering might be incorrect.
As answer give an array $[[x\_lower, y\_lower], [x\_upper, y\_upper]]$ with the coordinates of each cluster, following the format:

\begin{verbatim}
```json
{
  '<cluster index>': [[<lower bound x>, <lower bound y>],
    [<upper bound y>, <upper bound x>]],
}
```
\end{verbatim}
Your response should contain only the json data.
\end{tcolorbox}
\caption{Prompt for identifying the areas occupied by clusters.}
    \label{fig:clusters_area}
\end{figure}

\subsection{Series}

\begin{figure}[H]
    \centering
    \begin{tcolorbox}[colback=lightgray, colframe=darkgray, title=approximate.md, width=\textwidth]
    Please, approximate the plot with a series of points. Use up to 10 points. If you cannot provide the exact answer, give your best guess.
    The points, when connected, should approximate the plot. The points should cover the whole plot. The points have to be sorted by x value.

    Please respond in the following JSON format:

    \begin{verbatim}
    ```json
    {
        "points": [[<x1>, <y1>], [<x2>, <y2>], ...]
    }
    ```
    \end{verbatim}
    Your response should contain only the json data.
    \end{tcolorbox}
    \caption{Prompt for approximating a plot with a series of points.}
    \label{fig:approximate-prompt}
\end{figure}

\begin{figure}[H]
    \centering
    \begin{tcolorbox}[colback=lightgray, colframe=darkgray, title=min\_max\_interval.md, width=\textwidth]
    Please, tell me for which x values the y value is the highest and lowest. 
    Respond with intervals.

    Please respond in the following JSON format:

    \begin{verbatim}
    ```json
    {
        "max": [<lower bound for which x the y value is the highest>, 
            <upper bound for highest y value>],
        
        "min": [<lower bound for which x the y value is the lowest>, 
            <upper bound for which x the y value is the lowest>]
    }
    ```
    \end{verbatim}
    Your response should contain only the json data.
    \end{tcolorbox}
    \caption{Prompt for identifying intervals of highest and lowest y values.}
    \label{fig:min-max-interval}
\end{figure}

\begin{figure}[H]
    \centering
\begin{tcolorbox}[colback=lightgray, colframe=darkgray, title=missing\_data.md]
You are given a time series plot.

Consecutive data points are measured at equal intervals of time. However, sometimes the data is missing.

Please, help us with detecting whether the plot contains missing data or not.

Your answer should follow a format:

\begin{verbatim}
```json
{
    "missing_data": [x_1, x_2]
}
```
\end{verbatim}

Where `x\_1` and `x\_2` are the beginning and the end of the missing data interval. If the plot does not contain missing data, the answer should be:

\begin{verbatim}
```json
{}
```
\end{verbatim}
Your answer should contain only the json data. Do not include any other text or comments.
\end{tcolorbox}
\caption{Prompt for detecting missing data in a time series plot.}
    \label{fig:missing-data}
\end{figure}

\begin{figure}[H]
    \centering
\begin{tcolorbox}[colback=lightgray, colframe=darkgray, title=pointwise\_anomalies.md]
Please, help me detecting anomalies within the data. If you cannot provide the exact answer, give your best guess.

You are given a time series plot. Detect pointwise anomalies, if there are any. An anomaly is a data point that abruptly deviates from the other points.

Please, give your answer in the following format:

\begin{verbatim}
```json
{
  "anomalies": [] // list of x coordinates of possible anomalies
}
```
\end{verbatim}
If there are no anomalies in the plot, the "anomalies" list should be empty.

Your response should contain only the json data.
\end{tcolorbox}
\caption{Prompt for detecting pointwise anomalies in a time series plot.}
    \label{fig:pointwise-anomalies}
\end{figure}

\newpage
\subsection{Histograms}
\hspace{1em}

\begin{figure}[H]
    \centering
\begin{tcolorbox}[colback=lightgray, colframe=darkgray, title=min\_max\_bins.md]
Please specify the bin ranges in the histogram where the maximum and minimum values occur. If you're uncertain, please provide your best estimate of these bin ranges.Please respond in the following JSON format:

\begin{verbatim}
```json
{
    "min": [[<x1>, <x2>], [<x3>, <x4>], ...],
    "max": [[<x1>, <x2>], [<x3>, <x4>], ...],
}
```
\end{verbatim}
Your response should contain only the json data.
\end{tcolorbox}
\caption{Prompt for specifying bin ranges with maximum and minimum values in a histogram.}
    \label{fig:min_max_bins}
\end{figure}

\begin{figure}[H]
    \centering
\begin{tcolorbox}[colback=lightgray, colframe=darkgray, title=monotonicity.md]
Please identify two specific intervals in the histogram where values are constantly increasing and two where they are constantly decreasing. If you're unsure, please give your best estimate of these ranges.

Please respond in the following JSON format:

\begin{verbatim}
```json
{
    "increasing": [[<x1>, <x2>], [<x3>, <x4>]],
    "decreasing": [[<x1>, <x2>], [<x3>, <x4>],]
}
```
\end{verbatim}
Your response should contain only the json data.
\end{tcolorbox}
\caption{Prompt for identifying monotonic intervals in a histogram.}
    \label{fig:monotonicity}
\end{figure}

\begin{figure}[H]
    \centering
\begin{tcolorbox}[colback=lightgray, colframe=darkgray, title=anomalies.md]
Please specify the anomalies range. If you cannot provide the exact answer, give your best guess.

An anomaly in this context is either a missing bin (a bin that is expected but has zero count) or a bin that is outside the expected range of the data.
Please, give your answer in the following format:

\begin{verbatim}
```json
{
  "anomalies_range": [<x1>, <x2>]
}
```
\end{verbatim}

If there are no anomalies in the plot, the "anomalies\_range" list should be empty.

Your response should contain only the json data.
\end{tcolorbox}
\caption{Prompt for specifying the anomalies range in a histogram.}
    \label{fig:anomalies}
\end{figure}

\begin{figure}[H]
    \centering
\begin{tcolorbox}[colback=lightgray, colframe=darkgray, title=below\_x\_value\_percent.md]
Please identify what percentage of the histogram's data have values below \_\_value\_\_ on the x-axis. If you're unsure, please give your best estimate.

Please respond in the following JSON format:

\begin{verbatim}
```json
{
    "percentage_below": "<your answer in percent>"
}
```
\end{verbatim}

Your response should contain only the json data.
\end{tcolorbox}
\caption{Prompt for specifying the percentage of histogram data below a certain x-value.}
    \label{fig:below_x_value_percent}
\end{figure}

\begin{figure}[H]
    \centering
\begin{tcolorbox}[colback=lightgray, colframe=darkgray, title=distributions.md]
I need your help determining the type of distribution that best fits this data. Could you please identify the distribution from this list of options:

- UNIFORM

- NORMAL

- EXPONENTIAL

- POISSON

- MULTIMODAL

- SKEW\_LEFT

- SKEW\_RIGHT

Please respond in the following JSON format:

\begin{verbatim}
```json
{
  "distribution": "<distribution_name>"
}
```
\end{verbatim}
\end{tcolorbox}
\caption{Prompt for determining the type of distribution that best fits the data.}
    \label{fig:distributions}
\end{figure}

\subsection{Boxplots}

\begin{figure}[H]
    \centering
\begin{tcolorbox}[colback=lightgray, colframe=darkgray, title=iqr\_ranges.md]
Please specify the x values of the box plots that has the biggest and the smallest interquartile ranges. If you cannot provide the exact answer, give your best guess.

Please, give your answer in the following format:

\begin{verbatim}
```json
{
  "biggest_iqr_x": <x1>,
  "smallest_iqr_x": <x2>,
}
```
\end{verbatim}

Where \(\langle x1 \rangle\) and \(\langle x2 \rangle\) are the x values of the box plots with the biggest and smallest interquartile ranges, respectively. Your response should contain only the json data.
\end{tcolorbox}
\caption{Prompt for specifying x values with the biggest and smallest interquartile ranges in box plots.}
    \label{fig:iqr_ranges_boxplots}
\end{figure}

\begin{figure}[H]
    \centering
\begin{tcolorbox}[colback=lightgray, colframe=darkgray, title=medians.md]
Please specify the x values of the box plots that has the highest and the lowest median. If you cannot provide the exact answer, give your best guess.

Please, give your answer in the following format:

\begin{verbatim}
```json
{
  "highest_median_x": <x1>,
  "lowest_median_x": <x2>,
}
```
\end{verbatim}

Where \(\langle x1 \rangle\) and \(\langle x2 \rangle\) are the x values of the box plots with the highest and lowest medians, respectively.
Your response should contain only the json data.
\end{tcolorbox}
\caption{Prompt for specifying x values with the highest and lowest median in box plots.}
    \label{fig:medians_prompt_boxplots}
\end{figure}

\begin{figure}[H]
    \centering
\begin{tcolorbox}[colback=lightgray, colframe=darkgray, title=overall\_ranges.md]
Please specify the x values of the box plots that has the biggest and the smallest overall ranges. If you cannot provide the exact answer, give your best guess.

Please, give your answer in the following format:

\begin{verbatim}
```json
{
  "biggest_range_x": <x1>,
  "smallest_range_x": <x2>,
}
```
\end{verbatim}
Where \(\langle x1 \rangle\) and \(\langle x2 \rangle\) are the x values of the box plots with the biggest and smallest overall ranges, respectively.
Your response should contain only the json data.
\end{tcolorbox}
\caption{Prompt for specifying x values with the biggest and smallest overall ranges in box plots.}
    \label{fig:overall_ranges_boxplots}
\end{figure}

\subsection{Violin plots}

\begin{figure}[H]
    \centering
\begin{tcolorbox}[colback=lightgray, colframe=darkgray, title=iqr\_ranges.md]
Please specify the x values of the violin plots that has the biggest and the smallest interquartile ranges. If you cannot provide the exact answer, give your best guess.

Please, give your answer in the following format:

\begin{verbatim}
```json
{
  "biggest_iqr_x": <x1>,
  "smallest_iqr_x": <x2>,
}
```
\end{verbatim}

Where \(\langle x1 \rangle\) and \(\langle x2 \rangle\) are the x values of the violin plots with the biggest and smallest interquartile ranges, respectively. Your response should contain only the json data.
\end{tcolorbox}
\caption{Prompt for specifying x values with the biggest and smallest interquartile ranges in violin plots.}
    \label{fig:iqr_ranges}
\end{figure}

\begin{figure}[H]
    \centering
\begin{tcolorbox}[colback=lightgray, colframe=darkgray, title=medians.md]
Please specify the x values of the violin plots that has the highest and the lowest median. If you cannot provide the exact answer, give your best guess.

Please, give your answer in the following format:

\begin{verbatim}
```json
{
  "highest_median_x": <x1>,
  "lowest_median_x": <x2>,
}
```
\end{verbatim}

Where \(\langle x1 \rangle\) and \(\langle x2 \rangle\) are the x values of the violin plots with the highest and lowest medians, respectively.
Your response should contain only the json data.
\end{tcolorbox}
\caption{Prompt for specifying x values with the highest and lowest median in violin plots.}
    \label{fig:medians_prompt}
\end{figure}

\begin{figure}[H]
    \centering
\begin{tcolorbox}[colback=lightgray, colframe=darkgray, title=overall\_ranges.md]
Please specify the x values of the violin plots that has the biggest and the smallest overall ranges. If you cannot provide the exact answer, give your best guess.

Please, give your answer in the following format:

\begin{verbatim}
```json
{
  "biggest_range_x": <x1>,
  "smallest_range_x": <x2>,
}
```
\end{verbatim}
Where \(\langle x1 \rangle\) and \(\langle x2 \rangle\) are the x values of the violin plots with the biggest and smallest overall ranges, respectively.
Your response should contain only the json data.
\end{tcolorbox}
\caption{Prompt for specifying x values with the biggest and smallest overall ranges in violin plots.}
    \label{fig:overall_ranges}
\end{figure}

\subsection{Evaluation}

\begin{figure}[h]
  \begin{subfigure}[t]{0.3\textwidth}
    \centering
    \includegraphics[width=\textwidth]{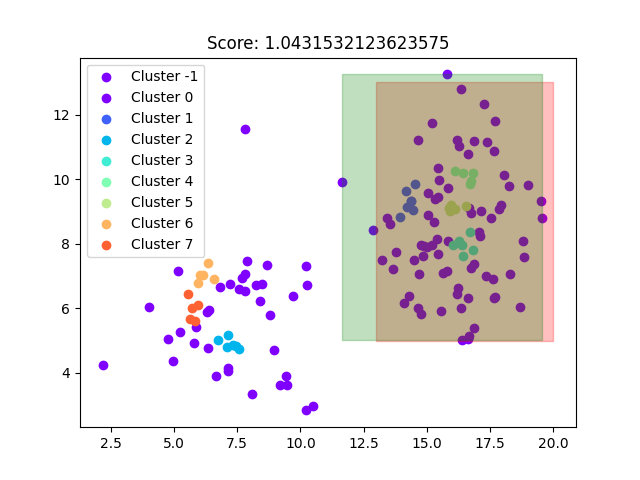}
    \caption{Visualization of the largest cluster identified by Claude-3-5-Sonnet.}
    \label{fig:biggest_cluster_sonnet}
  \end{subfigure}
    \hfill
      \centering
  \begin{subfigure}[t]{0.3\textwidth}
    \centering
    \includegraphics[width=\textwidth]{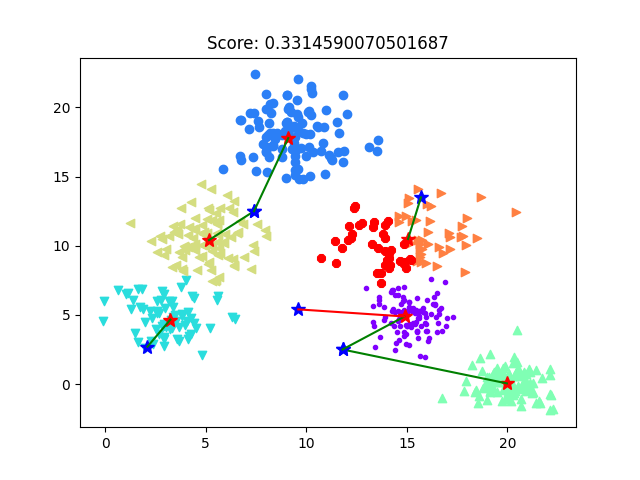}
    \caption{Centroids of clusters determined by Claude-3-Haiku.}
    \label{fig:cluster_centers_haiku}
  \end{subfigure}
  \hfill
  \begin{subfigure}[t]{0.3\textwidth}
    \centering
    \includegraphics[width=\textwidth]{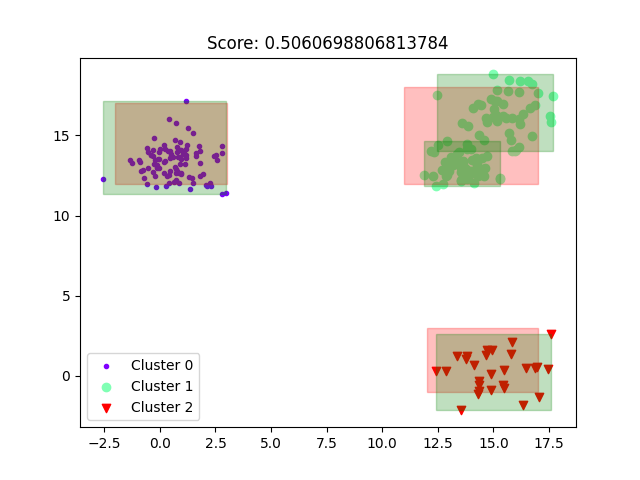} 
    \caption{Areas occupied by different clusters as identified by Gemini-1.5-Flash.}
    \label{fig:cluster_areas_gemini}
  \end{subfigure}
  \caption{Examples of clustering evaluation results using different models. Each subfigure shows specific aspects of cluster analysis: (a) the largest cluster, (b) cluster centroids, and (c) cluster areas.}
  \label{fig:clustering_evaluation}
\end{figure}


\newpage
\setcounter{table}{1}

\part{Extended Experimental Results}

\section{Experiment Setup}

\subsection{Tested Models}
In this study, we evaluate the performance of six different MLLMs on a series of visual interpretation tasks. The models tested include:

\paragraph{GPT-4o}
GPT-4o is an advanced language model in the GPT series, known for its robust text generation capabilities. It supports multimodality, allowing it to process both text and images. This enables the model to perform tasks such as image captioning, visual question answering, and generating text-based responses from visual inputs.

\paragraph{GPT-4o-mini}
GPT-4o-mini is a scaled-down version of GPT-4o, designed to offer similar multimodal capabilities but with fewer parameters and computational requirements. While it maintains the ability to handle text and images, its performance may be slightly limited compared to its larger counterpart.

\paragraph{Claude-3.5-Sonnet}
Claude-3.5-Sonnet is a variant of the Claude-3 series, optimized for creative and poetic text generation. Its multimodal capabilities are less emphasized, focusing primarily on text-based outputs. It excels in tasks that require creative language use but may not support extensive visual inputs.

\paragraph{Claude-3-Opus}
Claude-3-Opus is a more versatile model within the Claude-3 series, offering balanced multimodal capabilities. It can handle both text and images, making it suitable for applications in content creation that require the integration of visual elements with text.

\paragraph{Claude-3-Haiku}
Claude-3-Haiku is a specialized variant of the Claude-3 series, focused on generating short, concise poetic forms like haikus. Its multimodal capabilities are limited, with a primary emphasis on text-based creativity rather than integrating visual information.

\paragraph{Gemini-1.5-Pro}
Gemini-1.5-Pro is a highly capable multimodal model that excels in processing and integrating text and image inputs. It is designed for professional applications where high accuracy and seamless interaction between modalities are required, such as in detailed image analysis and text-based description generation.

\paragraph{Gemini-1.5-Flash}
Gemini-1.5-Flash is a faster, more streamlined version of Gemini-1.5-Pro, designed for quick multimodal processing with a focus on speed over detailed analysis. It can handle text and image inputs efficiently, making it suitable for real-time applications where quick response is critical.

\subsection{Querying process}

To ensure the reliability and reproducibility of the results, each model was queried three times on each dataset. This approach allows us to assess the consistency of the models' performance across different runs and provides a measure of variability in their outputs.

The datasets used in the experiments are designed to challenge the models' ability to interpret and analyze various types of visual data. For each query, the models were provided with a multimodal prompt consisting of a textual question paired with an image, and they were required to respond with structured answers in JSON format.

By conducting multiple runs for each model on every dataset, we aim to minimize the impact of random variations and obtain a more robust evaluation of each model's interpretive capabilities. The results from these runs were averaged to provide a comprehensive comparison across models and tasks.

\section{Results}

This section presents the performance comparison of various multimodal models across different datasets and tasks, including random walks, geometric walks, missing data, pointwise anomalies, clusters, box plots, and histograms. The results are segmented into different subsections to provide detailed insights into each task.

\subsection{Clusters Dataset}

The models' ability to handle clusters is evaluated in Table~\ref{tab:clusters_results}:

\begin{itemize}
    \item \textbf{Claude-3-5-Sonnet} outperformed other models in identifying the largest cluster, with high scores on both regular (0.96) and augmented (0.89) datasets. However, there was a noticeable decline in performance when predicting cluster centers on augmented data.
    \item \textbf{GPT-4o} performed well in identifying the biggest clusters but showed a mixed performance in determining cluster centers and cluster areas, particularly on augmented datasets.
    \item \textbf{Gemini-1.5-Pro} showed strong and consistent performance across all tasks, maintaining its scores relatively well even on augmented datasets.
\end{itemize}

\subsection{Series Dataset}
\subsubsection{Random Walk and Geometric Walk Datasets}

The performance of the models on random walk and geometric walk datasets is summarized in Table~\ref{tab:series_results_walks}. Overall, the models demonstrated varying capabilities in handling these datasets:

\begin{itemize}
    \item \textbf{GPT-4o} performed consistently well on the regular datasets but showed a decline in performance with the augmented datasets, particularly in the "Approximate" task where it scored significantly lower (-0.88) on random walks.
    \item \textbf{Claude-3-5-Sonnet} exhibited strong performance on the regular dataset, particularly for "Min-Max Interval Approximate," but showed substantial degradation on augmented datasets, with extreme negative scores in certain tasks.
    \item \textbf{Gemini-1.5-Pro} and \textbf{Gemini-1.5-Flash} maintained high scores across both regular and augmented datasets, although they also experienced some decline in performance on the augmented datasets.
\end{itemize}

\subsubsection{Missing Data and Pointwise Anomalies Datasets}

Table~\ref{tab:series_results_anomalies} details the performance on missing data and pointwise anomalies datasets:

\begin{itemize}
    \item \textbf{GPT-4o} showed solid performance on pointwise anomalies but struggled with missing data, particularly on augmented datasets.
    \item \textbf{Claude-3-5-Sonnet} consistently performed well on pointwise anomalies, achieving near-perfect scores (0.97) on both regular and augmented datasets. However, it did not perform as well on missing data.
    \item \textbf{Gemini-1.5-Pro} maintained high performance on pointwise anomalies but exhibited some difficulties with missing data, especially on augmented datasets.
\end{itemize}

\subsection{Histogram Dataset}

Table~\ref{tab:histogram_results} summarizes the performance of the models on histogram data:

\begin{itemize}
    \item \textbf{GPT-4o} showed strong performance in evaluating distributions and monotonicity, achieving relatively high scores across both regular and augmented datasets.
    \item \textbf{Claude-3-5-Sonnet} and \textbf{Gemini-1.5-Pro} performed consistently, especially in assessing monotonicity and "Below X Value Percent," with only slight variations between regular and augmented datasets.
    \item \textbf{GPT-4o-mini} and \textbf{Claude-3-Haiku} exhibited lower performance overall, particularly struggling with tasks like "Min Max Bins."
\end{itemize}

\subsection{Box Plot Dataset}

The evaluation of models on box plot data is presented in Table~\ref{tab:boxplot_results}:

\begin{itemize}
    \item \textbf{Claude-3-5-Sonnet} delivered the best results in assessing medians, overall ranges, and IQR ranges on regular datasets, though its performance decreased when dealing with augmented data.
    \item \textbf{GPT-4o} maintained moderate performance, with slightly lower scores on augmented datasets, particularly in estimating IQR ranges.
    \item \textbf{Gemini-1.5-Pro} was among the top performers, especially in overall ranges, and managed to maintain stability across both regular and augmented datasets.
\end{itemize}

\subsection{Violins Dataset}

Table~\ref{tab:violins_results} provides a general comparison across various metrics for violins like medians, overall ranges, and IQR ranges:

\begin{itemize}
    \item \textbf{Claude-3-5-Sonnet} emerged as the most consistent performer across all tasks, though it faced some challenges with augmented datasets.
    \item \textbf{GPT-4o} showed balanced performance across most tasks, though it was outperformed by others in specific areas, particularly with augmented datasets.
    \item \textbf{Gemini-1.5-Pro} demonstrated high and stable performance, making it a strong contender across different metrics.
\end{itemize}

\subsection{Summary of Key Findings}

\begin{itemize}
    \item \textbf{Model Variability}: The results indicate significant variability in how different models handle regular versus augmented datasets, with most models showing a decline in performance on augmented data.
    \item \textbf{Task-Specific Performance}: Certain models excelled in specific tasks, such as anomaly detection or cluster identification, while others performed better in general visualization tasks like box plots and histograms.
    \item \textbf{Impact of Augmentation}: Data augmentation generally posed challenges for the models, leading to lower scores and revealing areas where the models might require further tuning or improvements in handling complex or altered data.
\end{itemize}

\begin{table}[H]
\centering
\caption{Comparison of models' performance across regular and augmented clusters dataset}
\begin{tabular}{|l|c|c|c|c|c|c|c|c|c|}
\hline
\multirow{2}*{\textbf{Model}} & \multicolumn{2}{c|}{\textbf{Biggest Cluster}} & \multicolumn{2}{c|}{\textbf{Centers}} & \multicolumn{2}{c|}{\textbf{Clusters Area}} & \multicolumn{2}{c|}{\textbf{Average Score}} \\
\cline{2-9}
       & \textbf{Regular} & \textbf{Augmented} & \textbf{Regular} & \textbf{Augmented} & \textbf{Regular} & \textbf{Augmented} & \textbf{Regular} & \textbf{Augmented} \\
\hline
gpt-4o              & 0.68 & 0.86 & 0.58 & 0.47 & 0.36 & 0.30 & 0.54 & 0.55 \\
gpt-4o-mini         & 0.61 & 0.66 & 0.45 & 0.41 & 0.26 & 0.21 & 0.44 & 0.42 \\
claude-3-5-sonnet   & \textbf{0.96} & 0.89 & \textbf{0.73} & \textbf{0.59} & \textbf{0.50} & \textbf{0.42} & \textbf{0.73} & \textbf{0.63} \\
claude-3-opus       & 0.66 & \textbf{1.03} & 0.25 & 0.17 & 0.17 & 0.05 & 0.41 & 0.58 \\
claude-3-haiku      & 0.69 & 0.66 & 0.20 & 0.04 & 0.18 & 0.15 & 0.44 & 0.34 \\
gemini-1.5-pro      & 0.75 & 0.76 & \textbf{0.73} & 0.58 & 0.48 & 0.40 & 0.68 & 0.60 \\
gemini-1.5-flash    & 0.65 & 0.67 & 0.70 & \textbf{0.59} & 0.42 & 0.36 & 0.59 & 0.54 \\
\hline
\end{tabular}
\label{tab:clusters_results}
\end{table}

\begin{table}[H]

\centering
\caption{Comparison of models' performance across regular and augmented series dataset}

\begin{threeparttable}
{
    \centering
    
    \subcaption{Evaluation of random walk and geometric walk datasets.}
    \label{tab:series_results_walks}
    
    \begin{tabular}{|l|c|c|c|c|c|c|c|c|}
    
    \hline
    \multirow{3}*{\textbf{Model}} & \multicolumn{4}{c|}{\textbf{Random Walk}} & \multicolumn{4}{c|}{\textbf{Geometric Walk}}  \\
    \cline{2-9}
                   & \multicolumn{2}{c|}{\textbf{Min Max Interval}} & \multicolumn{2}{c|}{\textbf{Approximate}} & \multicolumn{2}{c|}{\textbf{Min Max Interval}} & \multicolumn{2}{c|}{\textbf{Approximate}}\\
        \cline{2-9}
                   & \textbf{Regular} & \textbf{Augmented} & \textbf{Regular} & \textbf{Augmented} & \textbf{Regular} &\textbf{Augmented} & \textbf{Regular} &\textbf{Augmented} \\
    \hline
    gpt-4o          & 0.78 & 0.52 & 0.60 & \textbf{-0.88} & 0.86 & 0.74 & 0.63 & \textbf{0.38}  \\
    gpt-4o-mini     & 0.64 & 0.45 & 0.04 & -0.87 & 0.66 & 0.69 & -0.03 & -0.03 \\
    claude-3-5-sonnet & \textbf{0.95} & \textbf{0.59} & 0.83 & -3490.95 & 0.94 & \textbf{0.79} & 0.81 & 0.22 \\
    claude-3-opus & 0.51 & 0.02 & -0.54 & -101.67 & 0.70 & 0.40 & -0.10 & -37.69 \\
    claude-3-haiku & 0.54 & 0.22 & -0.40 & -57.40 & 0.72 & 0.45 & -0.11 & -2.62  \\
    gemini-1.5-pro & 0.94 & 0.51 & 0.86 & -296.71 & 0.94 & 0.77 & 0.82 & -0.69 \\
    gemini-1.5-flash & \textbf{0.95} & 0.50 & \textbf{0.90} & -514.77 & \textbf{0.95} & 0.78 & \textbf{0.86} & -0.85\\
    \hline
    
    \end{tabular}
}
\bigskip

    {
    \centering
    
    \subcaption{Evaluation of missing data and pointwise anomalies datasets on relevant questions and average score.}
    \label{tab:series_results_anomalies}
    \makebox[\linewidth]{%

    \begin{tabular}{|l|c|c|c|c|c|c|}
    
    \hline
    \multirow{3}*{\textbf{Model}} &  \multicolumn{2}{c|}{\textbf{Missing Data}} & \multicolumn{2}{c|}{\textbf{Pointwise Anomalies}} & \multicolumn{2}{c|}{\textbf{Average Score}} \\
    \cline{2-7}
                   & \multicolumn{2}{c|}{\textbf{Missing Data}} & \multicolumn{2}{c|}{\textbf{Anomalies}} & \multicolumn{2}{c|}{} \\
        \cline{2-7}
                    & \textbf{Regular} & \textbf{Augmented} & \textbf{Regular} & \textbf{Augmented} & \textbf{Regular} &\textbf{Augmented}\tnote{1}\\
    \hline
    gpt-4o          & \textbf{0.33} & 0.26 & 0.89 & 0.69 & 0.68 & 0.52 \\
    gpt-4o-mini     & 0.19 & 0.24 & 0.81 & 0.68 & 0.39 & 0.41 \\
    claude-3-5-sonnet &  - & \textbf{0.31} & \textbf{0.97} & \textbf{0.97} & 0.90 & \textbf{0.58} \\
    claude-3-opus   & 0.24 & 0.21 & -3.45 & -8.28 & -0.44 & -9.07 \\
    claude-3-haiku  & 0.16 & 0.19 & 0.37 & 0.80 & 0.21 & -0.19 \\
    gemini-1.5-pro  &   -  &   -  & \textbf{0.97} & -0.89 & 0.90 & -0.08 \\
    gemini-1.5-flash &  -  & 0.21 & 0.95 & 0.75 & \textbf{0.92} & 0.28 \\
    \hline
    
    \end{tabular}
        }
    }
\begin{tablenotes}
\item[1] average score for augmented datasets excludes evaluation of random walk on approximate question
\end{tablenotes}
\end{threeparttable}
\label{tab:series_results}

\end{table}

\begin{table}[H]
\centering
\caption{Comparison of models' performance across regular and augmented histogram dataset}
\resizebox{\textwidth}{!}{%
    \begin{tabular}{|l|c|c|c|c|c|c|c|c|c|c|}
    \hline
    \multirow{2}*{\textbf{Model}} & \multicolumn{2}{c|}{\textbf{Distributions}} & \multicolumn{2}{c|}{\textbf{Min Max Bins}} & \multicolumn{2}{c|}{\textbf{Monotonicity}} & \multicolumn{2}{c|}{\textbf{Below x Value Percent}} & \multicolumn{2}{c|}{\textbf{Average Score}} \\
    \cline{2-11}
        & \textbf{Regular} & \textbf{Augmented} & \textbf{Regular} & \textbf{Augmented} & \textbf{Regular} & \textbf{Augmented} & \textbf{Regular} & \textbf{Augmented} & \textbf{Regular} & \textbf{Augmented} \\
    \hline
    gpt-4o          & 0.60 & \textbf{0.64} & 0.21 & 0.19 & 0.50 & 0.57 & \textbf{0.91} & \textbf{0.88} & 0.56 & \textbf{0.57} \\
    gpt-4o-mini     & 0.47 & 0.49 & 0.13 & 0.12 & 0.19 & 0.19 & 0.87 & 0.81 & 0.42 & 0.40 \\
    claude-3-5-sonnet & 0.47 & 0.54 & 0.25 & 0.20 & 0.26 & 0.26 & \textbf{0.91} & 0.84 & 0.47 & 0.46 \\
    claude-3-opus    & 0.35 & 0.31 & 0.13 & 0.08 & 0.17 & 0.17 & 0.77 & 0.73 & 0.35 & 0.32 \\
    claude-3-haiku   & 0.39 & 0.46 & 0.14 & 0.09 & 0.23 & 0.22 & 0.63 & 0.57 & 0.34 & 0.33 \\
    gemini-1.5-pro   & \textbf{0.62} & 0.50 & \textbf{0.31} & \textbf{0.27} & \textbf{0.58} & \textbf{0.58} & 0.83 & 0.80 & \textbf{0.59} & 0.54 \\
    gemini-1.5-flash & 0.51 & 0.49 & 0.28 & 0.21 & 0.31 & 0.28 & 0.78 & 0.77 & 0.47 & 0.44 \\
    \hline
    \end{tabular}
}
\label{tab:histogram_results}
\end{table}

\begin{table}[H]
\centering
\caption{Comparison of models' performance across regular and augmented box plot dataset}
\begin{tabular}{|l|c|c|c|c|c|c|c|c|}
\hline
\multirow{2}*{\textbf{Model}} & \multicolumn{2}{c|}{\textbf{Medians}} & \multicolumn{2}{c|}{\textbf{Overall Ranges}} & \multicolumn{2}{c|}{\textbf{IQR Ranges}} & \multicolumn{2}{c|}{\textbf{Average Score}} \\
\cline{2-9}
               & \textbf{Regular} & \textbf{Augmented} & \textbf{Regular} & \textbf{Augmented} & \textbf{Regular} & \textbf{Augmented} & \textbf{Regular} & \textbf{Augmented} \\
\hline
gpt-4o         & 0.61 & 0.52 & 0.44 & 0.37 & 0.57 & 0.44 & 0.54 & 0.44 \\
gpt-4o-mini    & 0.33 & 0.29 & 0.35 & 0.27 & 0.45 & 0.36 & 0.38 & 0.31 \\
claude-3-5-sonnet & \textbf{0.77} & \textbf{0.60} & \textbf{0.57} & 0.40 & \textbf{0.70} & 0.58 & \textbf{0.68} & \textbf{0.53} \\
claude-3-haiku    & 0.37 & 0.33 & 0.31 & 0.10 & 0.37 & 0.24 & 0.36 & 0.25 \\
gemini-1.5-pro         & 0.75 & 0.55 & 0.54 & \textbf{0.46} & 0.68 & \textbf{0.59} & \textbf{0.68} & \textbf{0.53} \\
gemini-1.5-flash           & 0.68 & 0.54 & 0.45 & 0.41 & 0.55 & 0.45 & 0.56 & 0.47 \\
\hline
\end{tabular}
\label{tab:boxplot_results}
\end{table}

\begin{table}[H]
\centering
\caption{Comparison of models' performance across regular and augmented violins dataset}
\begin{tabular}{|l|c|c|c|c|c|c|c|c|}
\hline
\multirow{2}*{\textbf{Model}} & \multicolumn{2}{c|}{\textbf{Medians}} & \multicolumn{2}{c|}{\textbf{Overall Ranges}} & \multicolumn{2}{c|}{\textbf{IQR Ranges}} & \multicolumn{2}{c|}{\textbf{Average Score}} \\
\cline{2-9}
       & \textbf{Regular} & \textbf{Augmented} & \textbf{Regular} & \textbf{Augmented} & \textbf{Regular} & \textbf{Augmented} & \textbf{Regular} & \textbf{Augmented} \\
\hline
gpt-4o              & 0.42 & 0.33 & 0.65 & 0.54 & 0.47 & 0.43 & 0.51 & 0.43 \\
gpt-4o-mini         & 0.25 & 0.23 & 0.51 & 0.41 & 0.40 & 0.34 & 0.39 & 0.32 \\
claude-3-5-sonnet   & \textbf{0.66} & 0.45 & \textbf{0.75} & 0.68 & 0.51 & 0.42 & \textbf{0.64} & 0.52 \\
claude-3-haiku      & 0.27 & 0.17 & 0.40 & 0.37 & 0.31 & 0.27 & 0.32 & 0.25 \\
gemini-1.5-pro      & 0.45 & \textbf{0.51} & 0.72 & \textbf{0.72} & \textbf{0.55} & \textbf{0.54} & 0.57 & \textbf{0.58} \\
gemini-1.5-flash    & 0.40 & 0.38 & 0.65 & 0.53 & 0.45 & 0.42 & 0.50 & 0.44 \\
\hline
\end{tabular}
\label{tab:violins_results}
\end{table}

\section{Analysis of Clustering Features and Scores}

\subsection{Regular Clustering Scores}

This subsection discusses the impact of various clustering configurations on scores when no image degradations are applied. The table ~\ref{tab:regular_clustering_scores} presents the scores for different clustering configurations.

\begin{table}[h!]
    \centering
    \caption{Scores for Regular Clustering Configurations}
    \begin{tabular}{lcccccc}
        \toprule
        \textbf{Clustering Configuration} & \textbf{Score} \\
        \midrule
        2 Centers & 0.6645 \\
        7 Centers & 0.4173 \\
        4 Centers & 0.6319 \\
        3 Centers & 0.6497 \\
        6 Centers & 0.5063 \\
        5 Centers & 0.4630 \\
        DBSCAN Algorithm & 0.5254 \\
        One Cluster Algorithm & 0.5791 \\
        Mean Shift Algorithm & 0.5676 \\
        K-Means Algorithm & 0.5245 \\
        Algorithm Parameters: n\_clusters=6 & 0.5620 \\
        Algorithm Parameters: n\_clusters=7 & 0.5659 \\
        Algorithm Parameters: n\_clusters=2 & 0.3886 \\
        Algorithm Parameters: n\_clusters=5 & 0.5179 \\
        Algorithm Parameters: n\_clusters=4 & 0.6344 \\
        Algorithm Parameters: n\_clusters=3 & 0.4592 \\
        Unique Markers: False & 0.5419 \\
        Unique Markers: True & 0.5696 \\
        Legend: True & 0.4985 \\
        Legend: False & 0.6163 \\
        Fill Clusters: False & 0.5911 \\
        Fill Clusters: True & 0.5233 \\
        \bottomrule
    \end{tabular}
    \label{tab:regular_clustering_scores}
\end{table}

From the regular clustering configurations, it is observed that:

- Number of Centers: The 2-center configuration achieved the highest score (0.6645), followed by the 3-center (0.6497) and 4-center (0.6319) configurations, suggesting that fewer centers might lead to better clustering performance in this context.

- Clustering Algorithms: The One Cluster algorithm outperformed others with a score of 0.5791, while the DBSCAN algorithm had a slightly lower score of 0.5254.

- Additional Features: The inclusion of a legend negatively impacted the scores (0.4985 with legend vs. 0.6163 without). Filling clusters also had a mixed effect, with a slightly lower score when clusters were filled (0.5233).

\subsection{Clustering with Image Degradations}

This subsection focuses on the scores of clustering configurations applied to datasets with image degradations. The table `\ref{tab:image_degradations_clustering_scores} presents these scores.

\begin{table}[h!]
    \centering
    \caption{Scores for Clustering with Image Degradations}
    \begin{tabular}{lcccccc}
        \toprule
        \textbf{Clustering Configuration} & \textbf{Score} \\
        \midrule
        4 Centers & 0.6295 \\
        6 Centers & 0.4367 \\
        7 Centers & 0.3812 \\
        3 Centers & 0.6378 \\
        5 Centers & 0.4643 \\
        2 Centers & 0.6510 \\
        One Cluster Algorithm & 0.5213 \\
        Mean Shift Algorithm & 0.5668 \\
        K-Means Algorithm & 0.4515 \\
        DBSCAN Algorithm & 0.4780 \\
        Algorithm Parameters: n\_clusters=5 & 0.5410 \\
        Algorithm Parameters: n\_clusters=2 & 0.3718 \\
        Algorithm Parameters: n\_clusters=6 & 0.2909 \\
        Algorithm Parameters: n\_clusters=4 & 0.5028 \\
        Unique Markers: True & 0.6582 \\
        Unique Markers: False & 0.5302 \\
        Legend: True & 0.5275 \\
        Legend: False & 0.5853 \\
        Fill Clusters: True & 0.4668 \\
        Fill Clusters: False & 0.5181 \\
        Augment: Rotate Image & 0.5403 \\
        Augment: Add Visual Noise & 0.4972 \\
        Augment: Random Add Image & 0.5675 \\
        \bottomrule
    \end{tabular}
    \label{tab:image_degradations_clustering_scores}
\end{table}

When image degradations are introduced:
- Number of Centers: The 2-center configuration again performed the best, achieving a score of 0.6510, indicating robustness against image degradations. The 3-center configuration also performed well (0.6378).
- Clustering Algorithms: The Mean Shift algorithm outperformed the others with a score of 0.5668. K-Means performed the worst in this context, with a score of 0.4515.
- Augmentation Effects: Among the augmentation techniques, "Random Add Image" achieved the highest score (0.5675), followed by "Rotate Image" (0.5403). Adding visual noise had the least positive effect, yielding a score of 0.4972.

\section{Analysis of Series Data: Random Walk Features and Scores}

\subsection{Regular Series Data: Random Walk Scores}

\begin{table}[h!]
    \centering
    \caption{Scores for Regular Series Data}
    \begin{tabular}{lcc}
        \toprule
        \textbf{Plot Configuration} & \textbf{Score} \\
        \midrule
        Black Color & 0.5495 \\
        Blue Color & 0.6126 \\
        Red Color & 0.4997 \\
        Green Color & 0.5095 \\
        Grid: True & 0.5193 \\
        Grid: False & 0.5560 \\
        \bottomrule
    \end{tabular}
    \label{tab:regular_series_scores}
\end{table}

Based on the results for regular series data ~\ref{tab:regular_series_scores}:
\begin{itemize}
    \item Color Influence: The blue color plot configuration achieved the highest score of 0.6126, indicating that blue is the most effective color for plotting in this context. Black color came next with a score of 0.5495.
    \item Grid Effect: The presence of a grid (`Grid: True`) resulted in a slightly lower score (0.5193) compared to no grid (`Grid: False`), which had a score of 0.5560. This suggests that a grid might slightly hinder performance.
\end{itemize}

\subsection{Series Data with Image Degradations}

\begin{table}[h!]
    \centering
    \caption{Scores for Series Data with Image Degradations}
    \begin{tabular}{lcc}
        \toprule
        \textbf{Plot Configuration} & \textbf{Score} \\
        \midrule
        Black Color & -8.2668 \\
        Red Color & -4.8859 \\
        Blue Color & -1220.9514 \\
        Green Color & -6.3501 \\
        Grid: False & -15.4225 \\
        Grid: True & -731.6747 \\
        Augment: Rotate Image & -579.6513 \\
        Augment: Add Visual Noise & 0.4843 \\
        Augment: Random Add Image & 0.4494 \\
        \bottomrule
    \end{tabular}
    \label{tab:image_degradation_series_scores}
\end{table}

When image degradations are introduced ~\ref{tab:image_degradation_series_scores}:
\begin{itemize}
    \item Color Influence: The blue color plot configuration had the most extreme negative score (-1220.9514), suggesting that it performs poorly under image degradation conditions. Black color had the least negative impact among color configurations (-8.2668).
    \item Grid Effect: The absence of a grid (`Grid: False`) resulted in a slightly less negative score (-15.4225) compared to having a grid (`Grid: True`), which had a much lower score (-731.6747).
    \item Augmentation Effects: Among the augmentation techniques, "Add Visual Noise" (0.4843) and "Random Add Image" (0.4494) had positive scores, showing better performance compared to other configurations under image degradation.
\end{itemize}

\section{Analysis of Histograms Features and Scores}

\subsection{Regular Histogram Scores}

The table ~\ref{tab:regular_histogram_scores} summarizes the scores for different histogram trends and plot features when no image degradations are applied.
\begin{table}[h!]
    \centering
    \caption{Histogram Scores for Regular Configurations}
    \begin{tabular}{lc}
        \toprule
        \textbf{Configuration} & \textbf{Score} \\
        \midrule
        Trend Type: Normal & 0.4594 \\
        Trend Type: Uniform & 0.4228 \\
        Trend Type: Skew Right & 0.6051 \\
        Trend Type: Poisson & 0.3506 \\
        Trend Type: Exponential & 0.4528 \\
        Trend Type: Multimodal & 0.4007 \\
        Trend Type: Skew Left & 0.3790 \\
        Plot Color: Blue & 0.4886 \\
        Plot Color: Black & 0.4381 \\
        Plot Color: Green & 0.4421 \\
        Plot Color: Red & 0.4837 \\
        Plot Color: Orange & 0.4355 \\
        Grid: False & 0.4485 \\
        Grid: True & 0.4660 \\
        \bottomrule
    \end{tabular}
    \label{tab:regular_histogram_scores}
\end{table}

From the regular histogram configurations, we observe the following:
\begin{itemize}
    \item Trend Types: The Skew Right trend type achieved the highest score (0.6051), indicating it best fits the data compared to others. The Poisson trend type had the lowest score (0.3506).
    \item Plot Colors: The Blue plot color had the highest score (0.4886), while the Orange color had the lowest score (0.4355). This suggests that Blue may be more effective in visualizing data.
    \item Grid Presence: The presence of a grid (Grid True) had a slightly better score (0.4660) compared to no grid (Grid False) at 0.4485, indicating that grids might enhance the interpretability of the histogram.
\end{itemize}

\subsection{Histogram Scores with Image Degradations}

The table ~\ref{tab:image_degraded_histogram_scores} presents the scores for histogram trends and features with image degradations applied.

\begin{table}[h!]
    \centering
    \caption{Histogram Scores with Image Degradations}
    \begin{tabular}{lc}
        \toprule
        \textbf{Configuration} & \textbf{Score} \\
        \midrule
        Trend Type: Normal & 0.4457 \\
        Trend Type: Skew Right & 0.5621 \\
        Trend Type: Poisson & 0.3439 \\
        Trend Type: Uniform & 0.4522 \\
        Trend Type: Multimodal & 0.4104 \\
        Trend Type: Skew Left & 0.2718 \\
        Trend Type: Exponential & 0.5309 \\
        Plot Color: Blue & 0.4318 \\
        Plot Color: Green & 0.3892 \\
        Plot Color: Red & 0.5544 \\
        Plot Color: Orange & 0.4250 \\
        Plot Color: Black & 0.4176 \\
        Grid: False & 0.4255 \\
        Grid: True & 0.4636 \\
        Augmentation: Random Add Image & 0.4713 \\
        Augmentation: Add Visual Noise & 0.3811 \\
        Augmentation: Rotate Image & 0.4637 \\
        \bottomrule
    \end{tabular}
    \label{tab:image_degraded_histogram_scores}
\end{table}

When image degradations are applied, the following observations can be made:
\begin{itemize}
    \item Trend Types: The Skew Right trend type achieved the highest score (0.5621), indicating its effectiveness even with degraded images. The Skew Left trend type had the lowest score (0.2718).
    \item Plot Colors: The Red plot color had the highest score (0.5544), while the Green color had the lowest score (0.3892). This suggests that Red might provide better visualization under degradation conditions.
    \item Augmentation Effects: Among the augmentation techniques, "Random Add Image" yielded the highest score (0.4713), suggesting it has the most positive impact on histogram performance compared to "Add Visual Noise" and "Rotate Image."
\end{itemize}

\section{Analysis of Boxplots Features and Scores}

\subsection{Regular Boxplots Scores}

The analysis of regular boxplots configurations ~\ref{tab:regular_boxplot_scores} reveals the following patterns:
\begin{itemize}
    \item Color Effects: The orange color configuration achieved the highest score (0.5356), closely followed by the green color (0.5417) and blue color (0.5176). Red color had a slightly lower score (0.5201).
    \item Grid Presence: The absence of a grid yielded a higher score (0.5465) compared to the presence of a grid (0.5126).
    \item Number of Series: The 5-series configuration performed the best (0.6022), with the 7-series (0.5548) and 9-series (0.4599) configurations also showing varied performance.
    \item Data Generators: The mixed data generator achieved the highest score (0.6279), suggesting it is the most effective for the given clustering setup. The normal data generator also performed well (0.5751), while the exponential generator (0.4689) and log-normal generator (0.4963) scored lower.
\end{itemize}

\begin{table}[h!]
    \centering
    \caption{Boxplot Scores for Regular Configurations}
    \begin{tabular}{lcccccccccc}
        \toprule
        \textbf{Configuration} & \textbf{Score} \\
        \midrule
        Blue Color & 0.5176 \\
        Black Color & 0.5348 \\
        Orange Color & 0.5356 \\
        Green Color & 0.5417 \\
        Red Color & 0.5201 \\
        Grid: False & 0.5465 \\
        Grid: True & 0.5126 \\
        9 Series & 0.4599 \\
        7 Series & 0.5548 \\
        5 Series & 0.6022 \\
        8 Series & 0.4948 \\
        10 Series & 0.4195 \\
        Mixed Data Generator & 0.6279 \\
        Exponential Data Generator & 0.4689 \\
        Normal Data Generator & 0.5751 \\
        Log-Normal Data Generator & 0.4963 \\
        \bottomrule
    \end{tabular}
    \label{tab:regular_boxplot_scores}
\end{table}

\subsection{Boxplots with Image Degradations}
For boxplots with image degradations ~\ref{tab:image_degradations_boxplot_scores}, the following observations were made:
\begin{itemize}
    \item Color Effects: The orange color configuration had the highest score (0.4646), with the blue color (0.4629) and red color (0.4551) showing moderate performance. The black color had the lowest score (0.3707).
    \item Grid Presence: The absence of a grid achieved a slightly higher score (0.4492) compared to the presence of a grid (0.4096).
    \item Augmentation Effects: Among the augmentations, "Random Add Image" performed the best with a score of 0.5459, while "Rotate Image" showed the lowest performance (0.2550). Adding visual noise had a moderate effect (0.4974).
    \item Number of Series: The 6-series configuration performed the best (0.6215), while the 10-series configuration had the lowest score (0.2795).
    \item Data Generators: The normal data generator achieved a high score (0.5216), whereas the mixed data generator (0.3487) and exponential data generator (0.4531) performed less favorably.
\end{itemize}

\begin{table}[h!]
    \centering
    \caption{Boxplot Scores for Clustering with Image Degradations}
    \begin{tabular}{lcccccccccc}
        \toprule
        \textbf{Configuration} & \textbf{Score} \\
        \midrule
        Blue Color & 0.4629 \\
        Black Color & 0.3707 \\
        Orange Color & 0.4646 \\
        Red Color & 0.4551 \\
        Green Color & 0.4026 \\
        Grid: False & 0.4492 \\
        Grid: True & 0.4096 \\
        Add Visual Noise Augmentation & 0.4974 \\
        Rotate Image Augmentation & 0.2550 \\
        Random Add Image Augmentation & 0.5459 \\
        9 Series & 0.4016 \\
        7 Series & 0.4388 \\
        5 Series & 0.4595 \\
        8 Series & 0.3486 \\
        6 Series & 0.6215 \\
        10 Series & 0.2795 \\
        Mixed Data Generator & 0.3487 \\
        Exponential Data Generator & 0.4531 \\
        Normal Data Generator & 0.4132 \\
        Log-Normal Data Generator & 0.5216 \\
        \bottomrule
    \end{tabular}
    \label{tab:image_degradations_boxplot_scores}
\end{table}

\section{Analysis of Violin Plots Features and Scores}

\subsection{Regular Violin Plots Scores}

\begin{table}[h!]
    \centering
    \caption{Scores for Regular Violin Plots Configurations}
    \begin{tabular}{lcccccc}
        \toprule
        \textbf{Feature} & \textbf{Score} \\
        \midrule
        Plot Color: Green & 0.5119 \\
        Plot Color: Red & 0.4810 \\
        Plot Color: Black & 0.4898 \\
        Plot Color: Blue & 0.4973 \\
        Plot Color: Orange & 0.4883 \\
        Plot Grid: True & 0.5097 \\
        Plot Grid: False & 0.4814 \\
        Number of Series: 5 & 0.6118 \\
        Number of Series: 7 & 0.5157 \\
        Number of Series: 6 & 0.5485 \\
        Number of Series: 8 & 0.4896 \\
        Number of Series: 9 & 0.4376 \\
        Number of Series: 10 & 0.4059 \\
        Generator: Mixed Data & 0.5479 \\
        Generator: Beta Data & 0.3728 \\
        Generator: Exponential Data & 0.5857 \\
        Generator: Uniform Data & 0.4954 \\
        Generator: Triangular Data & 0.3625 \\
        Generator: Gamma Data & 0.6774 \\
        Generator: Normal Data & 0.5456 \\
        Generator: Log Normal Data & 0.4786 \\
        Generator: Weibull Data & 0.4865 \\
        Generator: Cauchy Data & 0.3636 \\
        \bottomrule
    \end{tabular}
    \label{tab:regular_violin_scores}
\end{table}

The analysis of regular violin plots scores ~\ref{tab:regular_violin_scores} reveals several key patterns:
\begin{itemize}
    \item Color Schemes: The green plot color achieved the highest score (0.5119), while the black plot color had a lower score (0.4898).
    \item Plot Grid: Scores were higher when the grid was present (0.5097) compared to when it was absent (0.4814).
    \item Number of Series: The highest score was achieved with 5 series (0.6118), while the performance decreased with more series, especially at 10 series (0.4059).
    \item Data Generators: The Gamma data generator performed the best with a score of 0.6774, whereas the Beta data generator had the lowest score (0.3728). The Exponential data generator also performed well (0.5857).
\end{itemize}

\subsection{Violin Plots with Image Degradations}

\begin{table}[h!]
    \centering
    \caption{Scores for Violin Plots with Image Degradations}
    \begin{tabular}{lcccccc}
        \toprule
        \textbf{Feature} & \textbf{Score} \\
        \midrule
        Plot Color: Red & 0.3750 \\
        Plot Color: Blue & 0.3934 \\
        Plot Color: Orange & 0.4103 \\
        Plot Color: Green & 0.4706 \\
        Plot Color: Black & 0.4641 \\
        Plot Grid: True & 0.4641 \\
        Plot Grid: False & 0.3849 \\
        Augment: Rotate Image & 0.2627 \\
        Augment: Random Add Image & 0.4658 \\
        Augment: Add Visual Noise & 0.5664 \\
        Number of Series: 7 & 0.3924 \\
        Number of Series: 8 & 0.4175 \\
        Number of Series: 5 & 0.5675 \\
        Number of Series: 6 & 0.4631 \\
        Number of Series: 10 & 0.3417 \\
        Number of Series: 9 & 0.3937 \\
        Generator: Beta Data & 0.1842 \\
        Generator: Exponential Data & 0.5296 \\
        Generator: Triangular Data & 0.3750 \\
        Generator: Gamma Data & 0.7315 \\
        Generator: Normal Data & 0.3479 \\
        Generator: Log Normal Data & 0.4444 \\
        Generator: Mixed Data & 0.5130 \\
        Generator: Uniform Data & 0.4471 \\
        Generator: Cauchy Data & 0.2895 \\
        Generator: Weibull Data & 0.3144 \\
        \bottomrule
    \end{tabular}
    \label{tab:degraded_violin_scores}
\end{table}

The analysis of scores with image degradations ~\ref{tab:degraded_violin_scores} reveals the following trends:
\begin{itemize}
    \item Color Schemes: The green plot color performed the best (0.4706), while red had the lowest score (0.3750).
    \item Plot Grid: The presence of a plot grid (0.4641) did not show a significant improvement over its absence (0.3849).
    \item Augmentation Techniques: The "Add Visual Noise" augmentation resulted in the highest score (0.5664), whereas "Rotate Image" had the lowest score (0.2627). This suggests that visual noise might be less detrimental to violin plots performance compared to other augmentations.
    \item Number of Series: A higher number of series generally resulted in lower scores, particularly with 10 series (0.3417) performing the worst.
    \item Data Generators: The Gamma data generator achieved the highest score (0.7315), while the Beta data generator had the lowest (0.1842), indicating a significant impact of data generation methods on performance.
\end{itemize}